%% file: tigr_journal.tex
\documentclass[journal]{IEEEtran}
\ifCLASSINFOpdf
\else
\fi

\usepackage[utf8]{inputenc} 
\usepackage[T1]{fontenc}    
\usepackage{hyperref}       
\usepackage{url}            
\usepackage{booktabs}       
\usepackage{amsfonts}       
\usepackage{nicefrac}       
\usepackage{microtype}      
\usepackage{xcolor}         
\usepackage{soul}           
\usepackage{multirow}
\usepackage{booktabs}

\usepackage{graphicx}
\usepackage[lofdepth,lotdepth]{subfig}

\usepackage{algorithm}
\usepackage[noend]{algpseudocode}

\usepackage{amsmath, bm}
\newcommand*{\norm}[1]{\vert\vert#1\vert\vert}
\newcommand*{\dkl}[2]{\text{D}_{\text{KL}}\left(#1\middle\|#2\right)}
\DeclareMathOperator*{\argmax}{arg\,max}

\usepackage[colorinlistoftodos,prependcaption,textsize=tiny]{todonotes}
\usepackage{lipsum}                     
\usepackage{xargs}                      
\newcommandx{\unsure}[2][1=]{\todo[linecolor=red,backgroundcolor=red!25,bordercolor=red,#1]{#2}}
\newcommandx{\change}[2][1=]{\todo[linecolor=blue,backgroundcolor=blue!25,bordercolor=blue,#1]{#2}}
\newcommandx{\info}[2][1=]{\todo[linecolor=green,backgroundcolor=green!25,bordercolor=green,#1]{#2}}
\newcommandx{\improvement}[2][1=]{\todo[linecolor=Plum,backgroundcolor=Plum!25,bordercolor=Plum,#1]{#2}}

\newcommand*{\abs}[1]{\vert#1\vert}

\newcommand*\circled[1]{\tikz[baseline=(char.base)]{
    \node[shape=circle,draw,inner sep=0.2pt] (char) {#1};}}

\input{graphs}

\begin{document}
%
\title{Meta-Reinforcement Learning in Non-Stationary and Non-Parametric Environments}
%
%
%

\author{Zhenshan~Bing,
        Lukas~Knak,
        Fabrice~O.~Morin,
        Kai~Huang,~\IEEEmembership{Member,~IEEE}
        and~Alois~Knoll,~\IEEEmembership{Senior~Member,~IEEE}
\thanks{Z. Bing, L. Knak, F. Morin, and A. Knoll are with the Department
of Informatics, Technical University of Munich, Germany.
E-mail: \{bing, morinf, knoll\}@in.tum.de, lukas.knak@tum.de}
\thanks{K. Huang is with the School of Data and Computer Science, Sun Yat-sen University, China.}
\thanks{Manuscript received April 19, 2005; revised August 26, 2015.}}

%
%

\markboth{Journal of \LaTeX\ Class Files,~Vol.~14, No.~8, August~2015}%
{Shell \MakeLowercase{\textit{et al.}}: Bare Demo of IEEEtran.cls for IEEE Journals}
%



\maketitle

\begin{abstract}
Recent state-of-the-art artificial agents lack the ability to adapt rapidly to new tasks, as they are trained exclusively for specific objectives and require massive amounts of interaction to learn new skills. 
Meta-reinforcement learning (meta-RL) addresses this challenge by leveraging knowledge learned from training tasks to perform well in previously unseen tasks. 
However, current meta-RL approaches limit themselves to narrow parametric and stationary task distributions, ignoring qualitative differences and non-stationary changes between tasks that occur in the real world. 
In this paper, we introduce TIGR, a \textbf{T}ask-\textbf{I}nference-based meta-RL algorithm using explicitly parameterized \textbf{G}aussian variational autoencoders (VAE) and gated \textbf{R}ecurrent units, designed for non-parametric and non-stationary environments. 
We employ a generative model involving a VAE to capture the multi-modality of the tasks. 
We decouple the policy training from the task-inference learning and efficiently train the inference mechanism on the basis of an unsupervised reconstruction objective.
We establish a zero-shot adaptation procedure to enable the agent to adapt to non-stationary task changes.
We provide a benchmark with qualitatively distinct tasks based on the \textit{half-cheetah} environment and demonstrate the superior performance of TIGR compared to state-of-the-art meta-RL approaches in terms of sample efficiency (3-10 times faster), asymptotic performance, and applicability in non-parametric and non-stationary environments with zero-shot adaptation.
Videos can be viewed at 
\href{https://videoviewsite.wixsite.com/tigr}{https://videoviewsite.wixsite.com/tigr}.
\end{abstract}


%
\IEEEpeerreviewmaketitle

\input{01_introduction}
\input{02_background}
\input{03_related_work}
\input{04_problem_statement}
\input{05_methodology}
\input{06_experiments}
\input{07_conclusion}
\appendices

\input{08_appendix}



\section*{Acknowledgment}

This project/research has received funding from the European
Union’s Horizon 2020 Framework Programme for Research and
Innovation under the Specific Grant Agreement No.945539 (Human Brain Project SGA3).

\ifCLASSOPTIONcaptionsoff
  \newpage
\fi



\bibliographystyle{plain}
\bibliography{tigr_journal}

%

%

%

\begin{IEEEbiography}[{\includegraphics[width=1in, height=1.25in, clip, keepaspectratio]{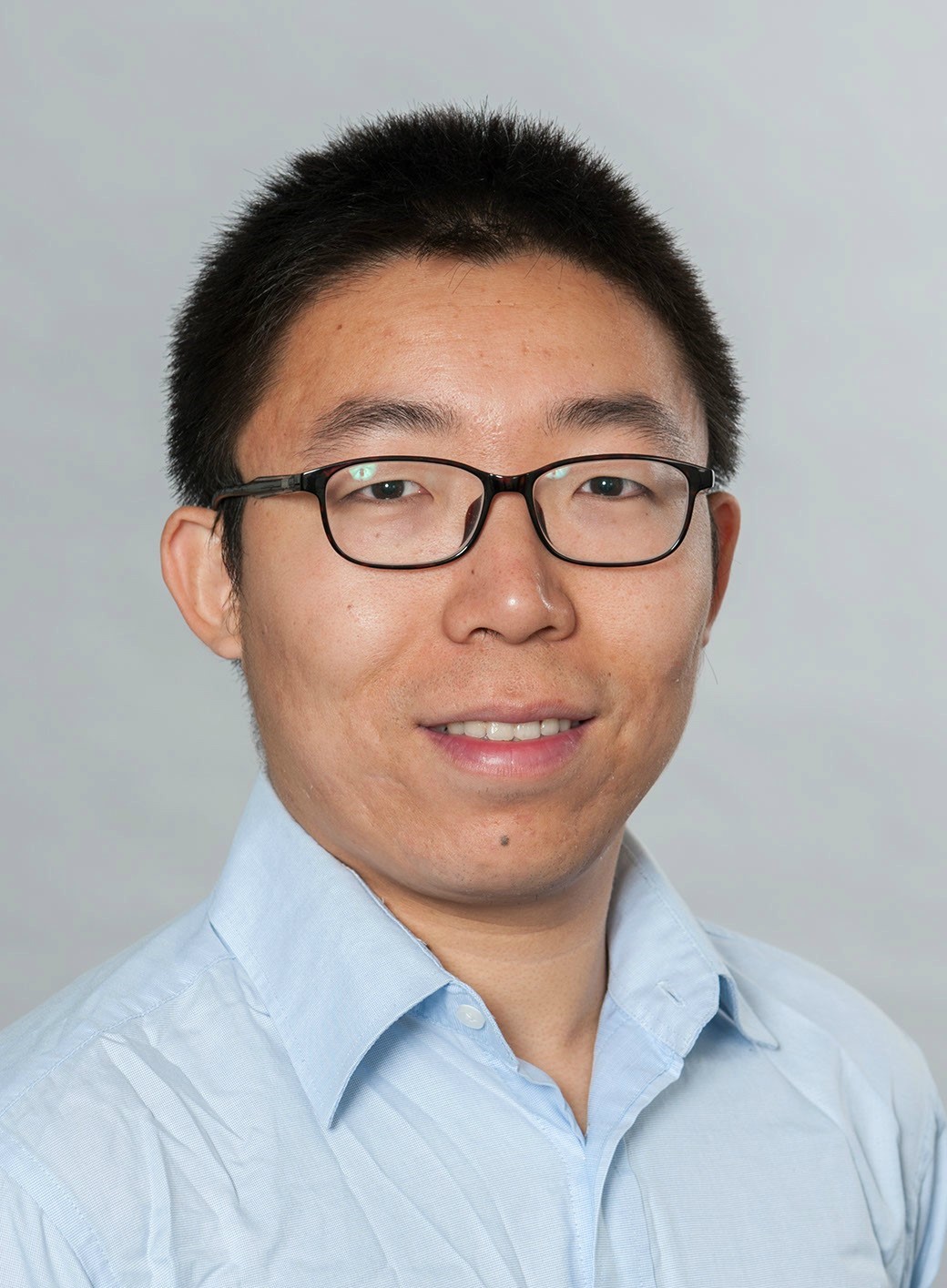}}]{Zhenshan Bing}
	received his doctorate degree in Computer Science from the Technical University of Munich, Germany, in 2019. 
	He received his B.S degree in Mechanical Design Manufacturing and Automation from Harbin Institute of Technology, China, in 2013,
	and his M.Eng degree in Mechanical Engineering in 2015, at the same university.
	Dr. Bing is currently a postdoctroal researcher with Informatics 6, Technical University of
	Munich, Germany.
	His research investigates bio-robots which are controlled by artificial neural networks and related applications.
\end{IEEEbiography}

\begin{IEEEbiographynophoto}{Lukas Knak}
    received his B.Sc degree in Cognitive Sciences from the University of Tübingen, Germany, in 2018,
	and his M.Sc. degree in Robotics, Cognition, Intelligence from the Technical University of
	Munich, Germany, in 2021.
	His interests include various applications of artificial intelligence in robotics.
\end{IEEEbiographynophoto}

\begin{IEEEbiography}[{\includegraphics[width=1in, height=1.25in, clip, keepaspectratio]{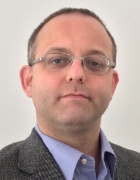}}]{Fabrice O. Morin}
	received an engineering degree from the Ecole Nationale Supérieure des Mines de Nancy (Nancy, France) in 1999, a Master's Degree in Bioengineering from the University of Strathclyde (Glasgow, United Kingdom) in 2000, and a Ph.D. in Materials Science from the Japanese Advanced Institute of Science and Technology (Nomi-Shi, Japan) in 2004.
	After several post-docs at the University of Tokyo (Japan) and the IMS laboratory (University of Bordeaux, France), in 2008 he joined Tecnalia, a nonprofit RTO in San Sebastián (Spain), first as a senior researcher, then as a group leader. 
	There, he worked on various projects in Neurotechnology and Biomaterials, funded both by public programs and private research contracts. 
	Since 2017, he has worked as a 
	scientific coordinator at the Technical University of Munich (Germany) where, in the framework of the Human Brain Project, he oversees the development of software tools for embodied simulation applied to Neuroscience and Artificial Intelligence.
\end{IEEEbiography}

\begin{IEEEbiography}[{\includegraphics[width=1in, height=1.25in, clip, keepaspectratio]{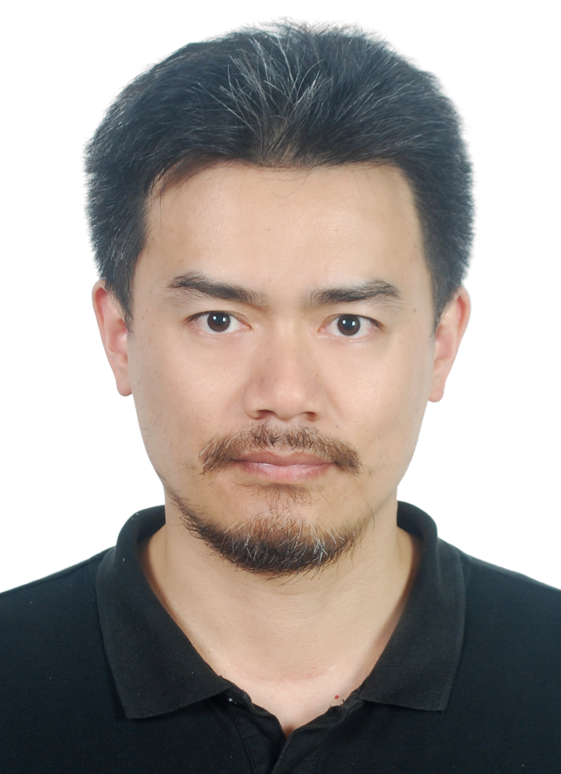}}]{Kai Huang}
	Kai Huang joined Sun Yat-Sen University as a Professor in 2015. He was appointed as the director of the Institute of Unmanned Systems of School of Data and Computer Science in 2016. He was a senior researcher in the Computer Science Department, the Technical University of Munich, Germany from 2012 to 2015 and a research group leader at fortiss GmbH in Munich, Germany, in 2011. He earned his Ph.D. degree at ETH Zurich, Switzerland, in 2010, his MSc from University of Leiden, the Netherlands, in 2005, and his BSc from Fudan University, China, in 1999. His research interests include techniques for the analysis, design, and optimization of embedded systems, particularly in the automotive and robotic domains. He was awarded the Program of Chinese Global Youth Experts 2014 and was granted the Chinese Government Award for Outstanding Self-Financed Students Abroad 2010. 
	He was the recipient of Best Paper Awards ESTC 2017, ESTIMedia 2013, SAMOS 2009, Best Paper Candidate ROBIO 2017, ESTMedia 2009, and General Chairs' Recognition Award for Interactive Papers in CDC 2009. He has served as a member of the technical committee on Cybernetics for Cyber-Physical Systems of IEEE SMC Society since 2015. 
\end{IEEEbiography}
\begin{IEEEbiography}[{\includegraphics[width=1in, height=1.25in, clip, keepaspectratio]{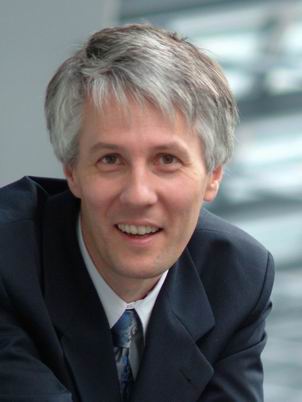}}]{Alois Knoll}
	(Senior Member) received his diploma
	(M.Sc.) degree in Electrical/Communications
	Engineering from the University of Stuttgart,
	Germany, in 1985 and his Ph.D. (\textit{summa cum
		laude}) in Computer Science from Technical
	University of Berlin, Germany, in 1988. He
	served on the faculty of the Computer Science
	department at TU Berlin until 1993. He joined the University of Bielefeld, Germany
	as a full professor and served as the director of the 
	Technical Informatics research group until 2001. Since 2001, he has been a
	professor at the Department of Informatics, Technical  University of Munich (TUM), Germany . He was also on the board of directors of
	the Central Institute of Medical Technology at TUM (IMETUM). From 2004 to 2006, he was Executive Director of the Institute of
	Computer Science at TUM. Between 2007 and 2009, he was a member
	of the EU's highest advisory board on information technology, ISTAG,
	the Information Society Technology Advisory Group, and a member
	of its subgroup on Future and Emerging Technologies (FET). In this
	capacity, he was actively involved in developing the concept of the EU's
	FET Flagship projects. His research interests include cognitive, medical
	and sensor-based robotics, multi-agent systems, data fusion, adaptive
	systems, multimedia information retrieval, model-driven development of
	embedded systems with applications to automotive software and electric
	transportation, as well as simulation systems for robotics and traffic.
	
\end{IEEEbiography}






\end{document}

%% file: graphs.tex
\definecolor{TUMBlue}{HTML}{0065BD}
\definecolor{TUMSecondaryBlue}{HTML}{005293}
\definecolor{TUMSecondaryBlue2}{HTML}{003359}
\definecolor{TUMBlack}{HTML}{000000}
\definecolor{TUMWhite}{HTML}{FFFFFF}
\definecolor{TUMDarkGray}{HTML}{333333}
\definecolor{TUMGray}{HTML}{808080}
\definecolor{TUMLightGray}{HTML}{CCCCC6}
\definecolor{TUMAccentGray}{HTML}{DAD7CB}
\definecolor{TUMAccentOrange}{HTML}{E37222}
\definecolor{TUMAccentGreen}{HTML}{A2AD00}
\definecolor{TUMAccentLightBlue}{HTML}{98C6EA}
\definecolor{TUMAccentBlue}{HTML}{64A0C8}

\usepackage{pgfplots}
\usepackage{pgfplotstable}
\usepackage{tikz}
\tikzset{>=stealth}
\usetikzlibrary{patterns}
\usetikzlibrary{pgfplots.statistics}
\usetikzlibrary{positioning, shapes}
\usetikzlibrary{calc}

\tikzstyle{neuron} = [circle, minimum size=0.5cm, text width=0.2cm, text height=0.2cm, text centered, draw=black, align=center, fill=TUMAccentGray, font=\footnotesize]
\tikzstyle{weight} = [thin,-,>=stealth, draw=black!80, align=center]

\tikzstyle{latent} = [rectangle, minimum width=1cm, minimum height=1cm, text centered, draw=black, rounded corners, align=center, fill=TUMAccentLightBlue]
\tikzstyle{loss} = [rectangle, minimum width=1cm, minimum height=1.2cm, text centered, draw=black, align=center]
\tikzstyle{module} = [rectangle, minimum width=2.5cm, minimum height=2cm, text centered, draw=black, rounded corners, align=center, fill=TUMBlue!10]
\tikzstyle{dark_module} = [rectangle, minimum width=2.5cm, minimum height=2cm, text centered, draw=black, rounded corners, align=center, fill=TUMAccentLightBlue]

\tikzstyle{arrow} = [thick,->,>=stealth,rounded corners=4pt, draw=black, align=center]
\tikzstyle{dashedarrow} = [thick,->,>=stealth,rounded corners=4pt, draw=black, align=center, dashed]
\tikzstyle{graydashedarrow} = [thick,->,>=stealth,rounded corners=4pt, draw=black, align=center, dashed, color=TUMGray]
\tikzstyle{orangedashedarrow} = [thick,->,>=stealth,rounded corners=4pt, draw=black, align=center, dashed, color=TUMAccentOrange]

\tikzstyle{marker} = [circle, minimum size=0.4cm, scale=0.8, text centered, draw=black, align=center, fill=orange!10, font=\large]

\newcommand{\drawpipeline}{
    
    \centering
    
    \begin{tikzpicture} [scale=0.75, transform shape]
        
        \node(context) [module, minimum width=5cm, minimum height=2cm, label={[align=center]Context $\bm c$ at timestep $t$}] at (-3, 0) {\begin{tabular}{cccc}
            $\bm s_{t-1}$ & $\bm a_{t-1}$ & $r_{t-1}$ & $\bm s_{t}$\\\hline
            $\bm s_{t-2}$ & $\bm a_{t-2}$ & $r_{t-2}$ & $\bm s_{t-1}$\\\hline
            & \multicolumn{2}{c}{$...$} & \\\hline
            $\bm s_{t-T}$ & $\bm a_{t-T}$ & $r_{t-T}$ & $\bm s_{t-T+1}$
        \end{tabular}};
        
        \node(over_encoder)[module, minimum width=6cm, minimum height=2cm, label={[align=center]Encoder \\(Task inference)}] at (3,0) {};
        
        \node(encoder) [dark_module, minimum width=2cm, minimum height=1.5cm, align=left] at (1.5,0) {GRU};
        \node(gmm) [dark_module, minimum width=2cm, minimum height=1.5cm] at (4.5,0) {VAE};

        \node(z) [latent, label={[align=center]Latent\\Variable}] at (7.2,0) {$\bm z$};
        
        \node(prediction) [module, minimum width=3cm, minimum height=1.5cm] at (10.5,2.5) {Decoder \\(State/Reward \\ prediction)};
        \node(sac) [module, minimum width=3cm, minimum height=1.5cm] at (10.5,0) {Task-conditioned \\policy (SAC)};
        
        \node(gradient_marker_1) [marker] at ($(prediction.west -| prediction.west) + (-0.4, 0.55)$) {1};

        \node(l_z) [loss, minimum width=1.3cm] at ($([yshift=2.5cm]gmm.north -| z.north) + (-0.4, 0)$) {\large{$\mathcal{L}_\text{KL-divergence}$}};
        \node(l_g) [loss, minimum width=2.4cm] at ($([yshift=2.5cm]gmm.north -| gmm.south) + (-0.1, 0)$) {\large{$\mathcal{L}_\text{euclid}$}\\ \large{$\mathcal{L}_\text{classification}$}};
        
        \node(l_p) [loss, minimum width=2.4cm] at ($([xshift=2cm]prediction.east)$) {\large{$\mathcal{L}_\text{dynamics}$}\\ \large{$\mathcal{L}_\text{rewards}$}};
        \node(l_s) [loss, minimum width=2.4cm] at ($([xshift=2cm]sac.east)$) {\large{$\mathcal{L}_\text{actor}$}\\ \large{$\mathcal{L}_\text{critic}$}};
        
		\draw[arrow] (context.east) -- (encoder.west);
		\draw[arrow] (encoder.east) -- (gmm.west);
		\draw[arrow] (gmm.east) -- (z.west);
		\draw[arrow] (z.east) -- (sac.west);
		\draw[arrow] (z.east) -- ([xshift=0.5cm]z.east) -- ([xshift=0.5cm]$(prediction.west -| z.east)$) -- (prediction.west);
		
		\draw[arrow] (prediction.east) -- (l_p.west);
		\draw[arrow] (sac.east) -- (l_s.west);
		
		\draw[arrow] ([yshift=0cm]gmm.north) -- ([xshift=0.1cm]l_g.south);
		\draw[arrow] ([yshift=0.8cm]z.north) -- ([xshift=0.4cm]l_z.south);
		
		\draw[orangedashedarrow, <-] ([yshift=0.2cm]encoder.east) -- ([yshift=0.2cm]gmm.west);
		\draw[orangedashedarrow, <-] ([yshift=0.2cm]gmm.east) -- ([yshift=0.2cm]z.west);
		
		\draw[orangedashedarrow, <-] ([yshift=0.2cm]z.east) -- ([xshift=0.3cm,yshift=0.2cm]z.east) -- ([xshift=0.3cm,yshift=0.2cm]$(prediction.west -| z.east)$) -- ([yshift=0.2cm]prediction.west);
		
		\draw[orangedashedarrow, <-] ([yshift=0.2cm]prediction.east) -- ([yshift=0.2cm]l_p.west);
		\draw[orangedashedarrow, <-] ([yshift=0.2cm]sac.east) -- ([yshift=0.2cm]l_s.west);
		
		\draw[orangedashedarrow, <-] ([xshift=0.2cm,yshift=0.0cm]gmm.north) -- ([xshift=0.3cm]l_g.south);
		\draw[orangedashedarrow, <-] ([xshift=0.2cm,yshift=0.8cm]z.north) -- ([xshift=0.6cm]l_z.south);
		
		\node(gradient_marker_2) [marker] at ($(l_s.west -| l_s.west) + (-0.4, 0.55)$) {2};

    \end{tikzpicture}
}

\newcommand{\drawgru}{
    
    \centering
    
    \begin{tikzpicture} [scale=0.6, transform shape]
        
        \node(context) [module, minimum width=5cm, minimum height=2.5cm, label={[align=center]Context $\bm c$ at Timestep $t$}] at (-6.5, 0) {\begin{tabular}{cccc}
            $\bm s_{t-1}$ & $\bm a_{t-1}$ & $r_{t-1}$ & $\bm s_{t}$\\\hline
            $\bm s_{t-2}$ & $\bm a_{t-2}$ & $r_{t-2}$ & $\bm s_{t-1}$\\\hline
            & \multicolumn{2}{c}{$...$} & \\\hline
            $\bm s_{t-T}$ & $\bm a_{t-T}$ & $r_{t-T}$ & $\bm s_{t-T+1}$
        \end{tabular}};
        
        \node(gru_mod) [module, minimum width=3.5cm, minimum height=2.5cm, label={[label distance=0.6cm]GRU}] at (0, 0) {};
        
        \node(m11) [neuron] at (-1.2,-0.7) {$\bm s$};
        \node(m12) [neuron] at (-0.4,-0.7) {$\bm a$};
        \node(m13) [neuron] at (0.4,-0.7) {$r$};
        \node(m14) [neuron] at (1.2,-0.7) {$\bm s'$};
        
        \node(m21) [neuron] at (-1.2,0.7) {};
        \node(m22) [neuron] at (-0.4,0.7) {};
        \node(m23) [neuron] at (0.4,0.7) {};
        \node(m24) [neuron] at (1.2,0.7) {};
        
        \node(features) [latent, minimum width=1cm, minimum height=4cm] at (4.0,0) {.\\.\\.\\.\\.\\.};
        
		\foreach \from in {m11,m12,m13,m14}
		    \foreach \to in {m21,m22,m23,m24}
                \draw [weight] (\from.north) to (\to.south);
        
		\draw[arrow] ($(context.east) + (0, -0.75)$) -- ($(context.east) + (0.5, -0.75)$) -- ($(context.east |- gru_mod.south) + (0.5, -0.5)$) -- ($(gru_mod.south) - (0, 0.5)$) -- ($(gru_mod.south) - (0, 0.2)$) -- ([yshift=-0.4cm]m11.south) -- (m11.south);
        \draw[arrow] ($(context.east) + (0, -0.25)$) -- ($(context.east) + (0.5, -0.25)$) -- ($(context.east |- gru_mod.south) + (0.5, -0.5)$) -- ($(gru_mod.south) - (0, 0.5)$) -- ($(gru_mod.south) - (0, 0.2)$) -- ([yshift=-0.4cm]m12.south)-- (m12.south);
        \draw[arrow] ($(context.east) + (0, 0.25)$) -- ($(context.east) + (0.5, 0.25)$) -- ($(context.east |- gru_mod.south) + (0.5, -0.5)$) -- ($(gru_mod.south) - (0, 0.5)$) -- ($(gru_mod.south) - (0, 0.2)$) -- ([yshift=-0.4cm]m13.south)-- (m13.south);
        \draw[arrow] ($(context.east) + (0, 0.75)$) -- ($(context.east) + (0.5, 0.75)$) -- ($(context.east |- gru_mod.south) + (0.5, -0.5)$) -- ($(gru_mod.south) - (0, 0.5)$) -- ($(gru_mod.south) - (0, 0.2)$) -- ([yshift=-0.4cm]m14.south)-- (m14.south);
        
		\draw[arrow, <-] ($(context.east) + (0.5, 0.8)$) -- ($(context.east |- gru_mod.north) + (0.5, 0.5)$) -- ($(gru_mod.north) + (0, 0.5)$) -- ($(gru_mod.north) + (0, 0.2)$) -- ([yshift=+0.4cm]m21.north) -- (m21.north);
        \draw[arrow, <-] ($(context.east) + (0.5, 0.8)$) -- ($(context.east |- gru_mod.north) + (0.5, 0.5)$) -- ($(gru_mod.north) + (0, 0.5)$) -- ($(gru_mod.north) + (0, 0.2)$) -- ([yshift=+0.4cm]m22.north)-- (m22.north);
        \draw[arrow, <-] ($(context.east) + (0.5, 0.8)$) -- ($(context.east |- gru_mod.north) + (0.5, 0.5)$) -- ($(gru_mod.north) + (0, 0.5)$) -- ($(gru_mod.north) + (0, 0.2)$) -- ([yshift=+0.4cm]m23.north)-- (m23.north);
        \draw[arrow, <-] ($(context.east) + (0.5, 0.8)$) -- ($(context.east |- gru_mod.north) + (0.5, 0.5)$) -- ($(gru_mod.north) + (0, 0.5)$) -- ($(gru_mod.north) + (0, 0.2)$) -- ([yshift=+0.4cm]m24.north)-- (m24.north);
        
        \draw[dashedarrow, -] ([xshift=0.3cm]gru_mod.west) node[rectangle,left,color=black,draw=black,fill=white]{GATES} -- ([xshift=-0.3cm]gru_mod.east);
        
        \draw[arrow] (gru_mod.east) -- (features.west) node[midway,above,color=black]{hidden\\state\\$\bm h$};
        
        \draw[graydashedarrow] (features.east) -- ([xshift=1cm]features.east);
        
        \draw[orangedashedarrow,<-] ([yshift=-0.2cm]gru_mod.east) -- ([yshift=-0.2cm]features.west);
        \draw[orangedashedarrow,<-] ([yshift=-0.2cm]features.east) -- ([xshift=1cm,yshift=-0.2cm]features.east);
        
    \end{tikzpicture}
}

\newcommand{\drawgmm}{
    
    \centering
    \begin{tikzpicture} [scale=0.75, transform shape]
        
        \node(features) [latent, label={[align=center]Input Features}, minimum width=1cm, minimum height=4cm] at (-4.5,0) {.\\.\\.\\.\\.\\.};
        
        \node(g_mod) [module, label={[align=center]Gaussian Components}, minimum width=5.5cm, minimum height=5cm] at (-0.2, 0) {};
        
        \node(g10) [neuron] at (-2.3,1.6) {};
        \node(g11) [neuron] at (-2.3,0.8) {};
        \node(g12) [neuron] at (-2.3,0) {};
        \node(g13) [neuron] at (-2.3,-0.8) {};
        \node(g14) [neuron] at (-2.3,-1.6) {};
        
        \node(nr_mod) [module, minimum width=3.5cm, minimum height=1.2cm] at (0.4, 1.6) {};
        \node(nm_mod) [module, minimum width=3.5cm, minimum height=1.2cm] at (0.4, 0.0) {};
        \node(ns_mod) [module, minimum width=3.5cm, minimum height=1.2cm] at (0.4, -1.6) {};
        
        \node(gr1) [neuron] at (-0.8,1.6) {$\rho_1$};
        \node(gr2) [neuron] at (0.0,1.6) {$\rho_2$};
        \node(gr3) [neuron] at (0.8,1.6) {...};
        \node(gr4) [neuron] at (1.6,1.6) {$\rho_k$};
        
        \node(gm1) [neuron] at (-0.8,0) {$\mu_1$};
        \node(gm2) [neuron] at (0.0,0) {$\mu_2$};
        \node(gm3) [neuron] at (0.8,0) {...};
        \node(gm4) [neuron] at (1.6,0) {$\mu_k$};
        
        \node(gs1) [neuron] at (-0.8,-1.6) {$\sigma_1^2$};
        \node(gs2) [neuron] at (-0.0,-1.6) {$\sigma_2^2$};
        \node(gs3) [neuron] at (0.8,-1.6) {...};
        \node(gs4) [neuron] at (1.6,-1.6) {$\sigma_k^2$};
        
        \node(zetas) [module, minimum width=1.4cm, minimum height=3.4cm, label=below:Samples] at (4.5, -0.8) {};
        
        \node(zeta1) [latent, minimum width=0.6cm, minimum height=0.6cm, text width=0.3cm, text height=0.3cm] at (4.5,0.4) {$\zeta_1$};
        \node(zeta2) [latent, minimum width=0.6cm, minimum height=0.6cm, text width=0.3cm, text height=0.3cm] at (4.5,-0.4) {$\zeta_2$};
        \node(zeta3) [latent, minimum width=0.6cm, minimum height=0.6cm, text width=0.3cm, text height=0.3cm] at (4.5,-1.2) {...};
        \node(zeta4) [latent, minimum width=0.6cm, minimum height=0.6cm, text width=0.3cm, text height=0.3cm] at (4.5,-2.0) {$\zeta_k$};
        
        \node(plz) [latent,fill=TUMWhite,minimum width=0.4] at (6.5,0) {$+$};
        \node(z) [latent, label={[align=center]Latent\\Variable}] at (8, 0) {$\bm z$};
        
        \node(l_z) [loss, minimum width=2.4cm] at ([yshift=-1.5cm]z.south) {$\mathcal{L}_\text{KL-divergence}$};
        \node(l_r) [loss, minimum width=2.4cm] at ([yshift=1.7cm]z.north) {$\mathcal{L}_\text{classification}$\\$\mathcal{L}_\text{euclid}$};
        
		\foreach \from in {g10,g11,g12,g13,g14}
		    \foreach \to in {nr_mod, nm_mod, ns_mod}
                \draw [weight] (\from.east) to (\to.west);
        
        \draw[graydashedarrow] ($(features.west) - (1.3, 0)$) -- (features.west);
        \draw[dashedarrow] (features.east) -- ($(g_mod.west) + (0,0.0)$);
        
        \draw[graydashedarrow] (nm_mod.east) -- ([xshift=0.5cm]nm_mod.east) -- ([xshift=-0.7cm]zeta2.west) -- (zeta2.west);
        \draw[graydashedarrow] (nm_mod.east) -- ([xshift=0.5cm]nm_mod.east) -- ([xshift=-0.7cm]zeta3.west) -- (zeta3.west);
        \draw[graydashedarrow] (nm_mod.east) -- ([xshift=0.5cm]nm_mod.east) -- ([xshift=-0.7cm]zeta4.west) -- (zeta4.west);
        \draw[dashedarrow] (nm_mod.east) -- ([xshift=0.5cm]nm_mod.east) -- ([xshift=-0.7cm]zeta1.west) -- (zeta1.west);
        
        \draw[graydashedarrow] (ns_mod.east) -- ([xshift=0.5cm]ns_mod.east) -- ([xshift=-0.7cm]zeta2.west) -- (zeta2.west);
        \draw[graydashedarrow] (ns_mod.east) -- ([xshift=0.5cm]ns_mod.east) -- ([xshift=-0.7cm]zeta3.west) -- (zeta3.west);
        \draw[graydashedarrow] (ns_mod.east) -- ([xshift=0.5cm]ns_mod.east) -- ([xshift=-0.7cm]zeta4.west) -- (zeta4.west);
        \draw[dashedarrow] (ns_mod.east) -- ([xshift=0.5cm]ns_mod.east) -- ([xshift=-0.7cm]zeta1.west) -- (zeta1.west);
        
        \draw[graydashedarrow] (nr_mod.east) -- ([xshift=3.2cm]nr_mod.east) -- ([xshift=-0.3cm,yshift=0.1cm]plz.west) -- ([yshift=0.1cm]plz.west);
        \draw[graydashedarrow] (nr_mod.east) -- ([xshift=3.2cm]nr_mod.east) -- ([xshift=-0.3cm,yshift=-0.1cm]plz.west) -- ([yshift=-0.1cm]plz.west);
        \draw[graydashedarrow] (nr_mod.east) -- ([xshift=3.2cm]nr_mod.east) -- ([xshift=-0.3cm,yshift=-0.3cm]plz.west) -- ([yshift=-0.3cm]plz.west);
        \draw[dashedarrow] (nr_mod.east) -- ([xshift=3.2cm]nr_mod.east) -- ([xshift=-0.3cm,yshift=0.3cm]plz.west) -- ([yshift=0.3cm]plz.west);
        
        \draw[graydashedarrow] (zeta2.east) -- ([xshift=0.7cm]zeta2.east) -- ([xshift=-0.3cm,yshift=0.1cm]plz.west) -- ([yshift=0.1cm]plz.west);
        \draw[graydashedarrow] (zeta3.east) -- ([xshift=0.7cm]zeta3.east) -- ([xshift=-0.3cm,yshift=-0.1cm]plz.west) -- ([yshift=-0.1cm]plz.west);
        \draw[graydashedarrow] (zeta4.east) -- ([xshift=0.7cm]zeta4.east) -- ([xshift=-0.3cm,yshift=-0.3cm]plz.west) -- ([yshift=-0.3cm]plz.west);
        \draw[dashedarrow] (zeta1.east) -- ([xshift=0.7cm]zeta1.east) -- ([xshift=-0.3cm,yshift=0.3cm]plz.west) -- ([yshift=0.3cm]plz.west);
        
        \draw[arrow] (plz.east) -- (z.west);
        
        \draw[graydashedarrow] (z.east) -- ([xshift=1.3cm]z.east);
        
        \draw[arrow] (z.south) -- (l_z.north);
        \draw[arrow] ($(g_mod.east |- l_r.west)$) -- (l_r.west);
        
        \draw[orangedashedarrow,<-] ([yshift=0.2cm]z.east) -- ([xshift=1.3cm,yshift=0.2cm]z.east);
        \draw[orangedashedarrow,<-] ([xshift=0.2cm]z.south) -- ([xshift=0.2cm]l_z.north);
        \draw[orangedashedarrow,<-] ($(g_mod.east |- l_r.west) + (0,0.2)$) -- ([yshift=0.2cm]l_r.west);
        
        \draw[orangedashedarrow,<-] ([yshift=0.2cm]plz.east) -- ([yshift=0.2cm]z.west);
        
        \draw[orangedashedarrow,<-] ([yshift=0.2cm]nr_mod.east) -- ([xshift=3.3cm,yshift=0.2cm]nr_mod.east) -- ([xshift=-0.2cm,yshift=0.4cm]plz.west) -- ([yshift=0.4cm]plz.west);
        
        \draw[orangedashedarrow,<-] ([xshift=0.4cm,yshift=-0.2cm]zeta4.east) -- ([xshift=0.8cm,yshift=-0.2cm]zeta4.east) -- ([xshift=-0.2cm,yshift=-0.4cm]plz.west) -- ([yshift=-0.4cm]plz.west);
        
        \draw[orangedashedarrow,<-] ([yshift=0.2cm]nm_mod.east) -- ([xshift=0.6cm,yshift=0.2cm]nm_mod.east) -- ([xshift=-0.6cm,yshift=0.2cm]zeta1.west) -- ([xshift=-0.5cm,yshift=0.2cm]zeta1.west);
        
        \draw[orangedashedarrow,<-] ([yshift=-0.2cm]ns_mod.east) -- ([xshift=0.4cm,yshift=-0.2cm]ns_mod.east) -- ([xshift=-0.6cm,yshift=-0.2cm]zeta4.west) -- ([xshift=-0.5cm,yshift=-0.2cm]zeta4.west);
        
        \draw[orangedashedarrow,<-] ($(features.west) - (1.3, 0.2)$) -- ($(features.west) - (0,0.2)$);
        \draw[orangedashedarrow,<-] ($(features.east) - (0,0.2)$) -- ($(g_mod.west) - (0,0.2)$);

    \end{tikzpicture}
}

\newcommand{\drawprediction}{
    
    \centering
    
    \begin{tikzpicture} [scale=0.8, transform shape]
        
        \node(z) [latent, label={[align=center]Task\\ Representation}] at (-2,0) {$\bm z$};
        
        \node(d_mod) [module, minimum width=3.5cm, label={[align=center]Dynamics Network}, minimum height=2.5cm] at (1.65, 1.8) {};
        
        \node(d_s) [neuron] at (1,2.6) {$\bm s$};
        \node(d_a) [neuron] at (1,1.8) {$\bm a$};
        \node(d_z) [neuron] at (1,1) {$\bm z$};
        
        \node(d_s2) [neuron] at (2,2.6) {};
        \node(d_a2) [neuron] at (2,1.8) {};
        \node(d_z2) [neuron] at (2,1) {};
        
        \node(d_s3) [neuron] at (3,1.8) {$\bm s'$};
        
        \node(r_mod) [module, minimum width=3.5cm, label={[align=center]Rewards Network}, minimum height=2.5cm] at (1.65, -1.8) {};
        
        \node(r_s) [neuron] at (1,-1) {$\bm s$};
        \node(r_a) [neuron] at (1,-1.8) {$\bm a$};
        \node(r_z) [neuron] at (1,-2.6) {$\bm z$};
        
        \node(r_s2) [neuron] at (2,-1) {};
        \node(r_a2) [neuron] at (2,-1.8) {};
        \node(r_z2) [neuron] at (2,-2.6) {};
        
        \node(r_r3) [neuron] at (3,-1.8) {$r$};
        
        \node(l_d) [loss] at (5.5, 1.8) {$\mathcal{L}_\text{dynamics}$};
        \node(l_r) [loss] at (5.5, -1.8) {$\mathcal{L}_\text{rewards}$};
        
		\foreach \from in {d_s,d_a,d_z}
		    \foreach \to in {d_s2,d_a2,d_z2}
                \draw [weight] (\from) to (\to);
                
		\foreach \from in {d_s2,d_a2,d_z2}
		    \foreach \to in {d_s3}
                \draw [weight] (\from.east) to (\to.west);
        
		\foreach \from in {r_s, r_a, r_z}
		    \foreach \to in {r_s2, r_a2, r_z2}
                \draw [weight] (\from) to (\to);
                
		\foreach \from in {r_s2, r_a2, r_z2}
		    \foreach \to in {r_r3}
                \draw [weight] (\from.east) to (\to.west);
        
        \draw[graydashedarrow] ($(z.west) - (1.3, 0)$) -- (z.west);
        
        \draw[arrow] (z.east) -- ($(z.east) + (1, 0)$) -- ($(d_z.west) - (0.7, 0)$) -- (d_z.west);
        \draw[arrow] (z.east) -- ($(z.east) + (1, 0)$) -- ($(r_z.west) - (0.7, 0)$) -- (r_z.west);
        
        \draw[graydashedarrow] ($(d_s.west) - (0.65, 0)$) -- (d_s.west);
        \draw[graydashedarrow] ($(d_a.west) - (0.65, 0)$) -- (d_a.west);
        
        \draw[graydashedarrow] ($(r_s.west) - (0.65, 0)$) -- (r_s.west);
        \draw[graydashedarrow] ($(r_a.west) - (0.65, 0)$) -- (r_a.west);
        
        \draw[arrow] (d_s3.east) -- (l_d.west);
        \draw[arrow] (r_r3.east) -- (l_r.west);
        
        \draw[orangedashedarrow] ([yshift=0.2cm]l_d.west) -- ([yshift=0.2cm,xshift=0.1cm]d_s3.east);
        \draw[orangedashedarrow] ([yshift=-0.2cm]l_r.west) -- ([yshift=-0.2cm,xshift=0.1cm]r_r3.east);
        
        \draw[orangedashedarrow] ($(d_z.west) - (0.8,-0.1)$) -- ($(d_z.west) - (1,-0.1)$) -- ($(z.east) + (0.8, 0.2)$) -- ([yshift=0.2cm]z.east);
        \draw[orangedashedarrow] ($(r_z.west) - (0.8,0.1)$) -- ($(r_z.west) - (1,0.1)$) -- ($(z.east) + (0.8, -0.2)$) -- ([yshift=-0.2cm]z.east);
        
        \draw[orangedashedarrow, <-] ($(z.west) - (1.3, 0.2)$) -- ([yshift=-0.2cm]z.west);
        
        \draw[graydashedarrow] ($(l_d.north) + (0,0.5)$) node[above,color=black]{$s'$} -- (l_d.north);
        \draw[graydashedarrow] ($(l_r.north) + (0,0.5)$) node[above,color=black]{$r$} -- (l_r.north);
        
    \end{tikzpicture}
}

%% file: 01_introduction.tex
\section{Introduction}


\IEEEPARstart{H}{umans} have the ability to learn new skills by transferring previously acquired knowledge, which enables them to quickly and easily adapt to new challenges.
However, state-of-the-art artificial agents lack this ability, since they are generally trained on specific tasks from scratch, which renders them unable to adapt to differing tasks or to reuse existing experiences.
For instance, to imbue a robotic hand with the dexterity
to solve a Rubik’s Cube, OpenAI reported a cumulative experience of thirteen thousand years \cite{rubiks}. 
In contrast, adult humans are able to manipulate the cube almost instantaneously, as they possess prior knowledge regarding generic object manipulation.



As a promising approach, meta-RL reinterprets this open challenge of adapting to new and yet related tasks as a learning-to-learn problem \cite{learningtolearn}.
Specifically, meta-RL aims to learn new skills by first learning a prior from a set of similar tasks and then reusing this policy to succeed after few or zero trials in the new target environment.
Recent studies in meta-RL can be divided into three main categories.
Gradient-based meta-RL approaches, such as MAML \cite{maml}, aim to learn a set of highly sensitive model parameters, so that the agent can quickly adapt to new tasks with only few gradient descent steps.
Recurrence-based methods aim to learn how to implicitly store task information in the hidden states during meta-training and utilize the resulting mechanism during meta-testing \cite{recurrentmetarl}. 
While these two concepts can adapt to new tasks in only a few trials, they adopt on-policy RL algorithms during meta-training, which require massive amounts of data and lead to sample inefficiency. 
To address this issue, PEARL \cite{pearl}, a model-free and off-policy method, achieves state-of-the-art results and significantly outperforms prior studies in terms of sample efficiency and asymptotic performance, by representing the task with a single Gaussian distribution through an encoder which outputs the probabilistic task embeddings.

However, most previous approaches, including PEARL, are severely limited to narrow task distributions, as they have only been applied to parametric environments \cite{pearl, graphneuralnetwork, maml, metarlasti, modelbasedmeta}, in which only certain parameters of the tasks are varied.
This ignores the fact that humans are usually faced with qualitatively different tasks in their daily lives, which happen to share some common structure.
For example, grasping a bottle and turning a doorknob both require the dexterity of a hand.
However, the non-parametric variability introduced by the two different objects makes it much more difficult to solve the tasks when compared to the sole use of parametric variations, such as turning a doorknob to different angles.
%
%
Additionally, algorithms as PEARL \cite{pearl} and MAML \cite{maml} are designed to adapt to the task in few trials, which excludes them from being applicable to non-stationary environments, where task changes can occur at any time during the interaction.
In spite of these limitations, there is currently no study that explicitly focuses on both non-parametric and non-stationary environments while providing the benefits of model-free and off-policy algorithms, such as the superior data efficiency and good asymptotic performance.

In this paper, we establish an approach that addresses the challenge of learning how to behave in non-parametric and broad task distributions with non-stationary task changes. 
We leverage insights from PEARL \cite{pearl} and introduce a \textbf{T}ask-\textbf{I}nference-based meta-RL algorithm using explicitly parameterized \textbf{G}aussian variational autoencoders and gated \textbf{R}ecurrent units (TIGR), which is sample-efficient, adapts in a zero-shot manner to non-stationary task changes, and achieves good asymptotic performance in non-parametric tasks. 
We propose a novel algorithm composed of four concepts.
First, we use a generative model, leveraging a combination of Gaussians to cluster the information on each qualitatively different base task. 
Second, we decouple the task-inference training from the RL algorithm by reconstructing the tasks' Markov decision processes (MDPs) in an unsupervised setup.
Third, we propose a zero-shot adaptation mechanism by extracting features from recent transition history and infer task information at each timestep to enable the agent to adapt to task changes at any time.
Last, we provide a benchmark with non-parametric tasks based on the commonly used \textit{half-cheetah} environment.
Experiment results demonstrate that TIGR significantly outperforms state-of-the-art methods with 3-10 times faster sample efficiency, substantially increased asymptotic performance, and unmatched task-inference capabilities under zero-shot adaptation in non-parametric and non-stationary environments for the first time.
To the best of the authors’ knowledge, TIGR is the first model-free meta-RL algorithm to solve non-parametric and non-stationary environments with zero-shot adaptation.

%% file: 02_background.tex
\section{Background}

\subsection{Meta-reinforcement learning}
The learning problem of meta-RL is extended to an agent that has to solve different tasks from a distribution $p(\mathcal{T})$ \cite{metaworld}. 
Each task $\mathcal{T}$ is defined as an individual MDP specifying its properties. A meta-RL agent is not given any task information other than the experience it gathers while interacting with the environment. 
A standard meta-RL setup consists of two task sets: a meta-training task set $\mathcal{D}^{train}_{\mathcal{T}}$ used to train the agent, and a meta-test task set $\mathcal{D}^{test}_{\mathcal{T}}$ used to evaluate the agent.
Both sets are drawn from the same distribution $p(\mathcal{T})$, but $\mathcal{D}^{test}_{\mathcal{T}}$ may differ from $\mathcal{D}^{train}_{\mathcal{T}}$. 
The objective is to train a policy $\pi_{\theta}$ on $\mathcal{D}^{train}_{\mathcal{T}}$ that maximizes rewards on $\mathcal{D}^{test}_{\mathcal{T}}$, which is defined as 
\begin{equation}
    \theta^* = \argmax_\theta \mathbb{E}_{\mathcal{T} \sim \mathcal{D}^{test}_{\mathcal{T}}} \left[ \mathbb{E}_{\tau \sim p(\tau|\pi_\theta)} \left[\sum_{t} \gamma^t r_t\right] \right] \text{.}
    \label{meta_rl_objective}
\end{equation}
\subsection{Meta-training and meta-testing}
Meta-RL consists of two stages: meta-training and meta-testing. 
During meta-training, each training epoch consists of a data collection and optimization phase.
In the data collection phase, interaction experiences for each task are collected and stored in the replay buffer.
In the optimization phase, the losses for the policy are computed and the gradient of the averaged losses is used to update the parameters of the policy.
During meta-testing, the policy is adapted to new tasks with either few trials, meaning that the agent can experience the presented environment and adapt before the final evaluation, or in zero-shot manner, which means that the agent must solve the environment at first sight.

\subsection{Stationary and non-stationary environments}
The meta-RL setting can involve stationary or non-stationary environments. In stationary environments, each episode consists of one task, i.e., the underlying MDP of the environment is fixed during an episode. In such cases, an algorithm can use a few-shot (episode-wise) mechanism to adapt to the task by collecting experiences for a few episodes and adjusting before the final evaluation in the last trial.
In non-stationary environments, on the other hand, the underlying MDP can change at any timestep. Here, episode-wise adaptation fails and a zero-shot procedure with online (or continuous) adaptation at the transition level is required. However, the environment must exhibit local consistency over a number of timesteps during which the task is fixed \cite{Nagabandi2018} so that the agent can process the information from recent transitions to behave according to the objective at that time. Thus, we consider non-stationary environments that consist of several stationary sub-tasks that the agent must adapt to. This is a particularly common scenario in the real world, for example, when a motor malfunction occurs in a robot's joint that changes its dynamics and therefore impacts the MDP underlying the task.

\subsection{Parametric and non-parametric variability in meta-RL}
\begin{figure*}[!t]
\centering
     \subfloat[][Run forward]{
      \includegraphics[width=0.18\textwidth]{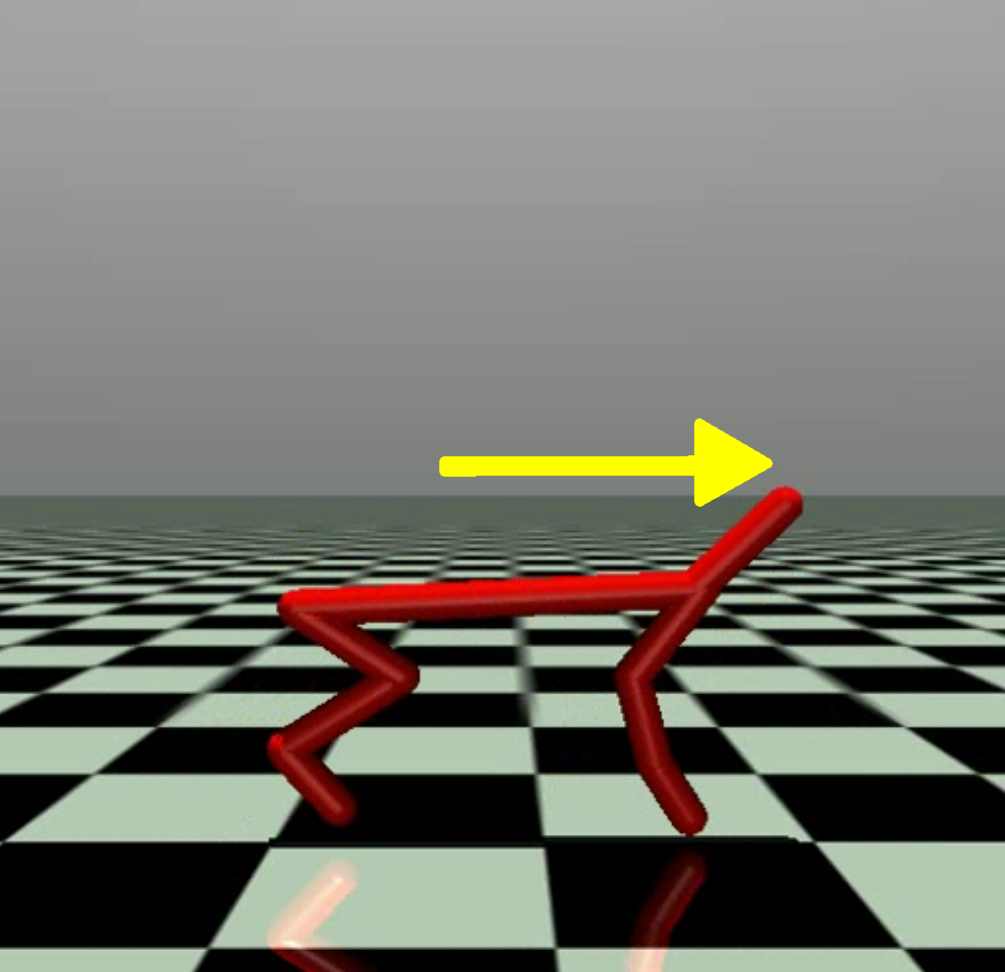}
     }
     \subfloat[][Run backward]{
      \includegraphics[width=0.18\textwidth]{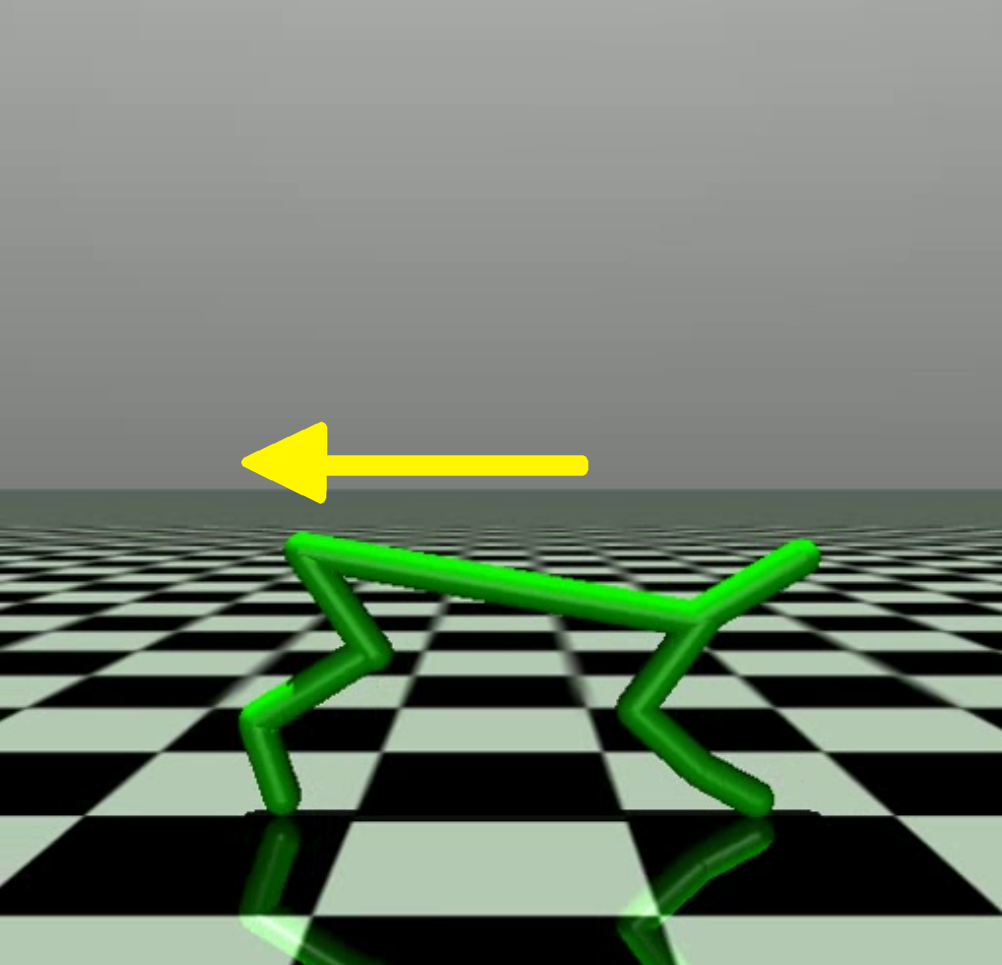}
     }
     \subfloat[][Reach front goal]{
      \includegraphics[width=0.18\textwidth]{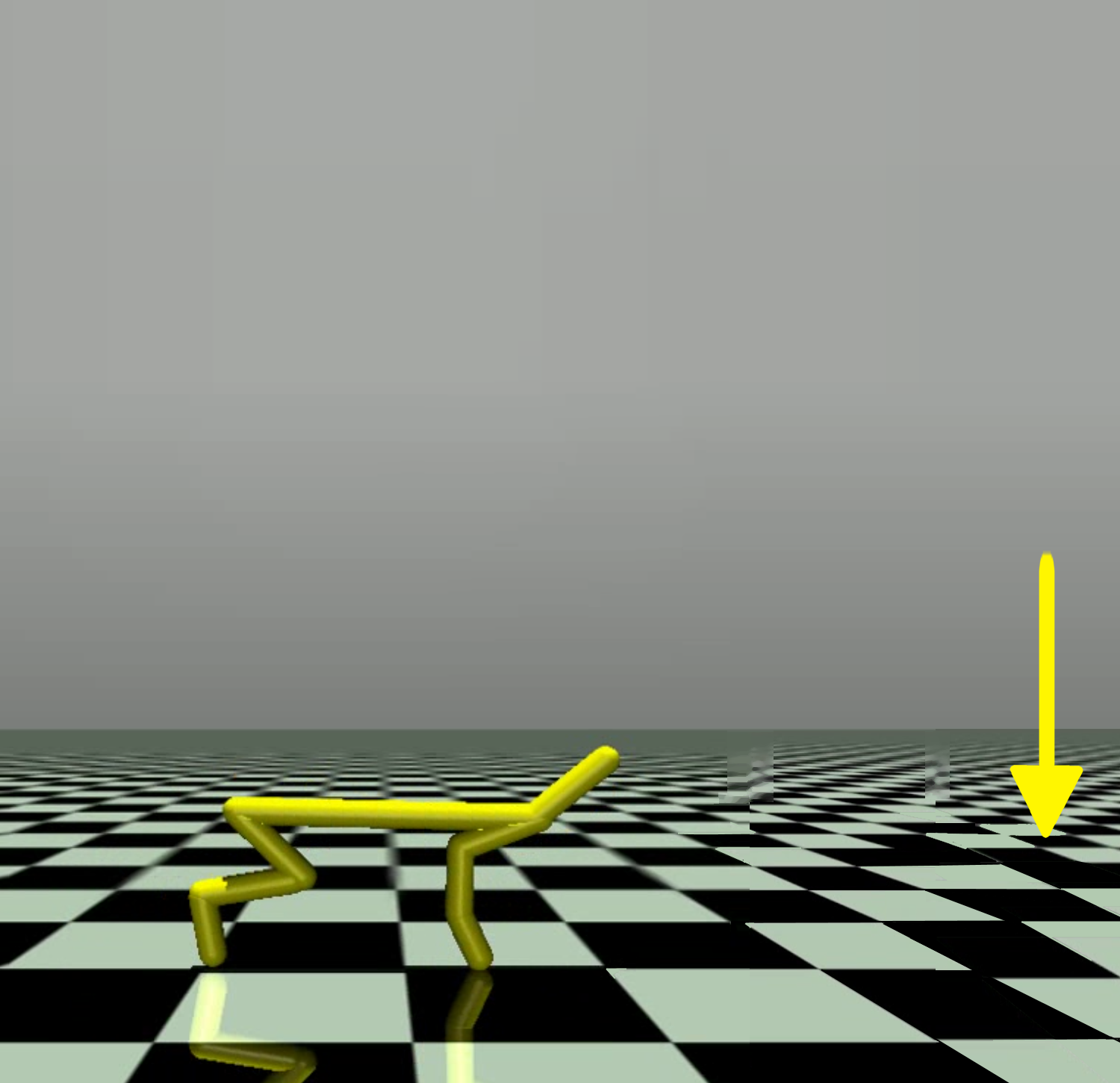}
     }
     \subfloat[][Reach back goal]{
      \includegraphics[width=0.18\textwidth]{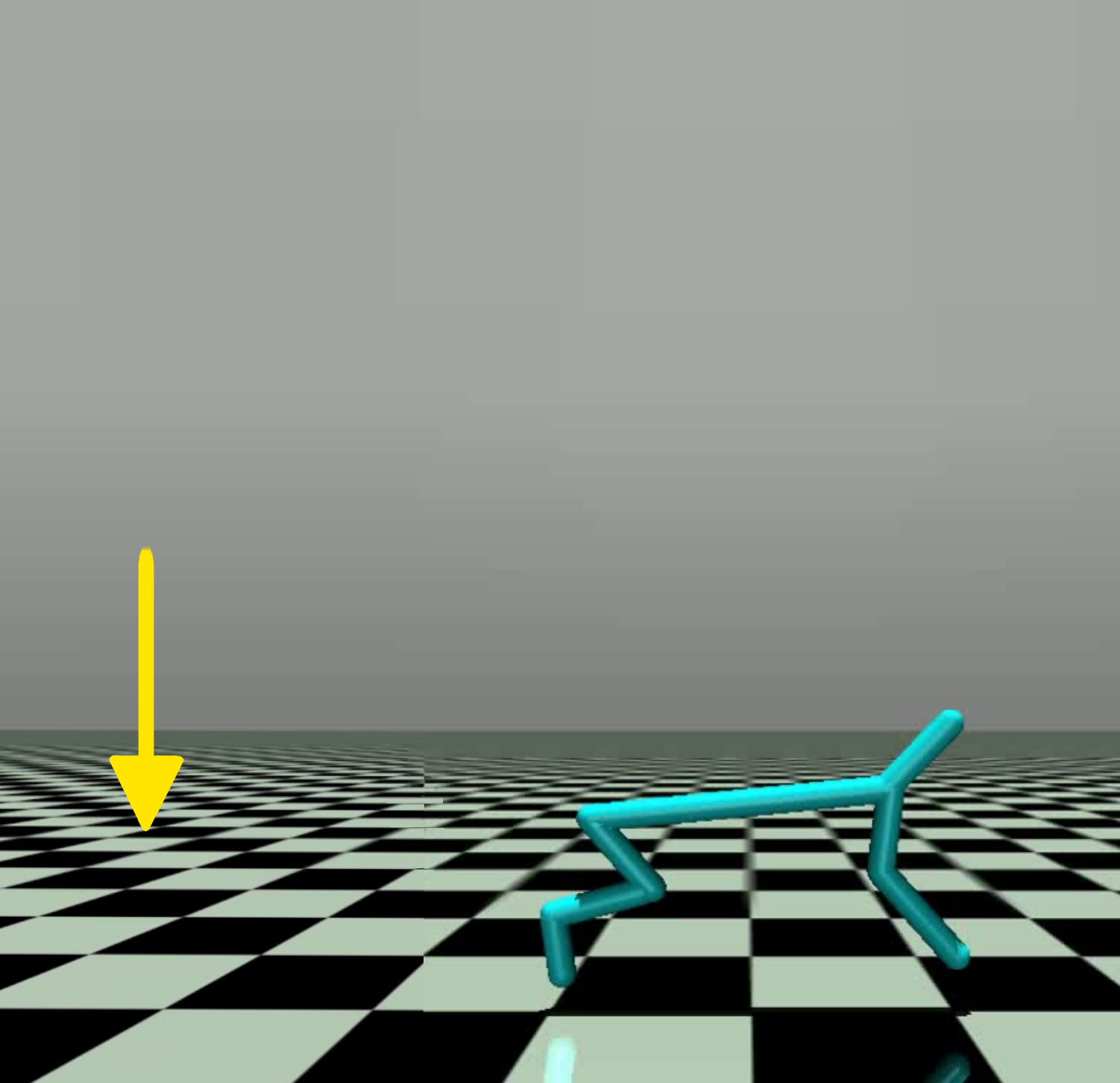}
     }\\
     \subfloat[][Front stand]{
      \includegraphics[width=0.18\textwidth]{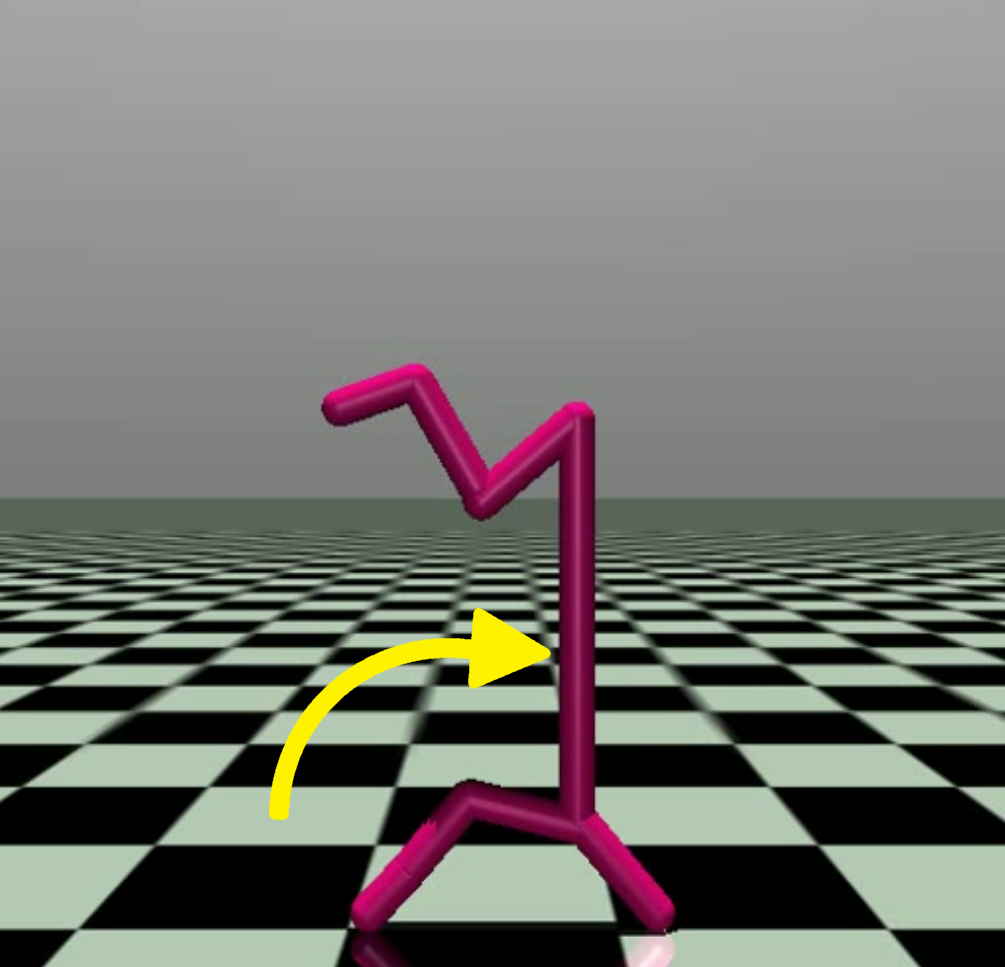}
     }
     \subfloat[][Back stand]{
      \includegraphics[width=0.18\textwidth]{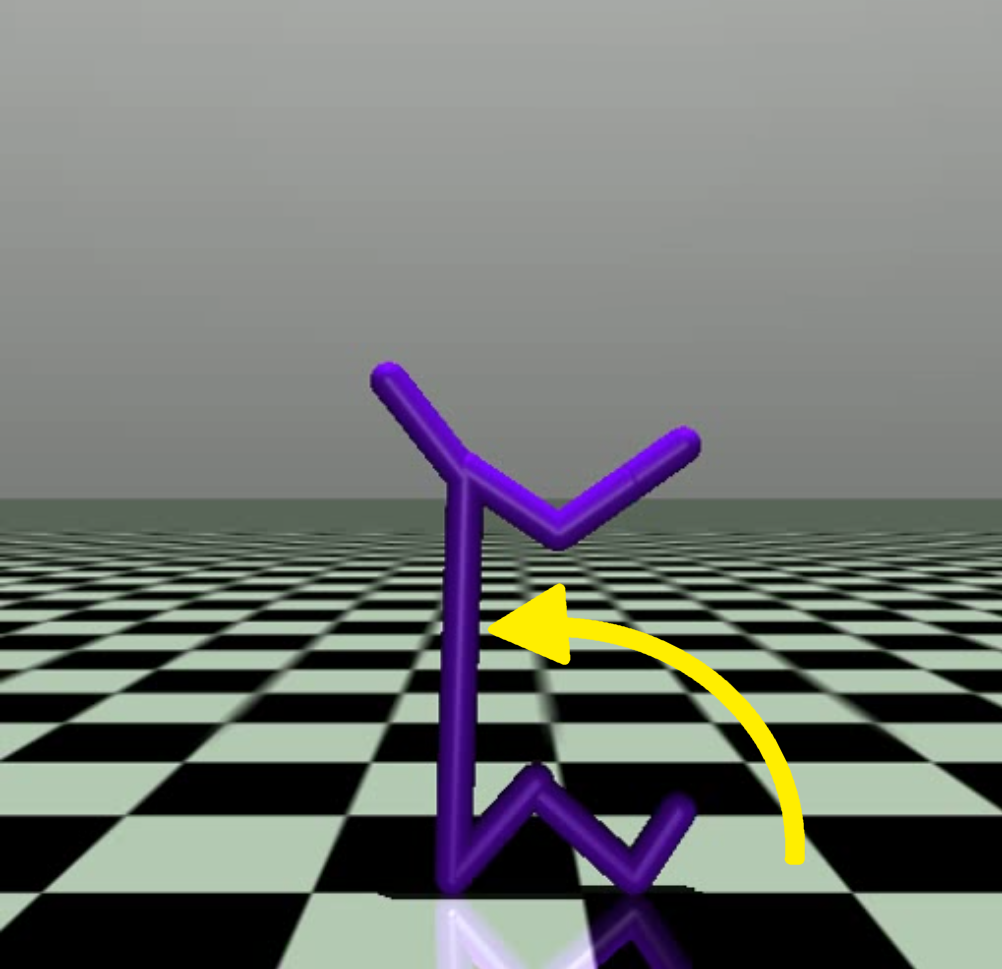}
     }
     \subfloat[][Jump]{
      \includegraphics[width=0.18\textwidth]{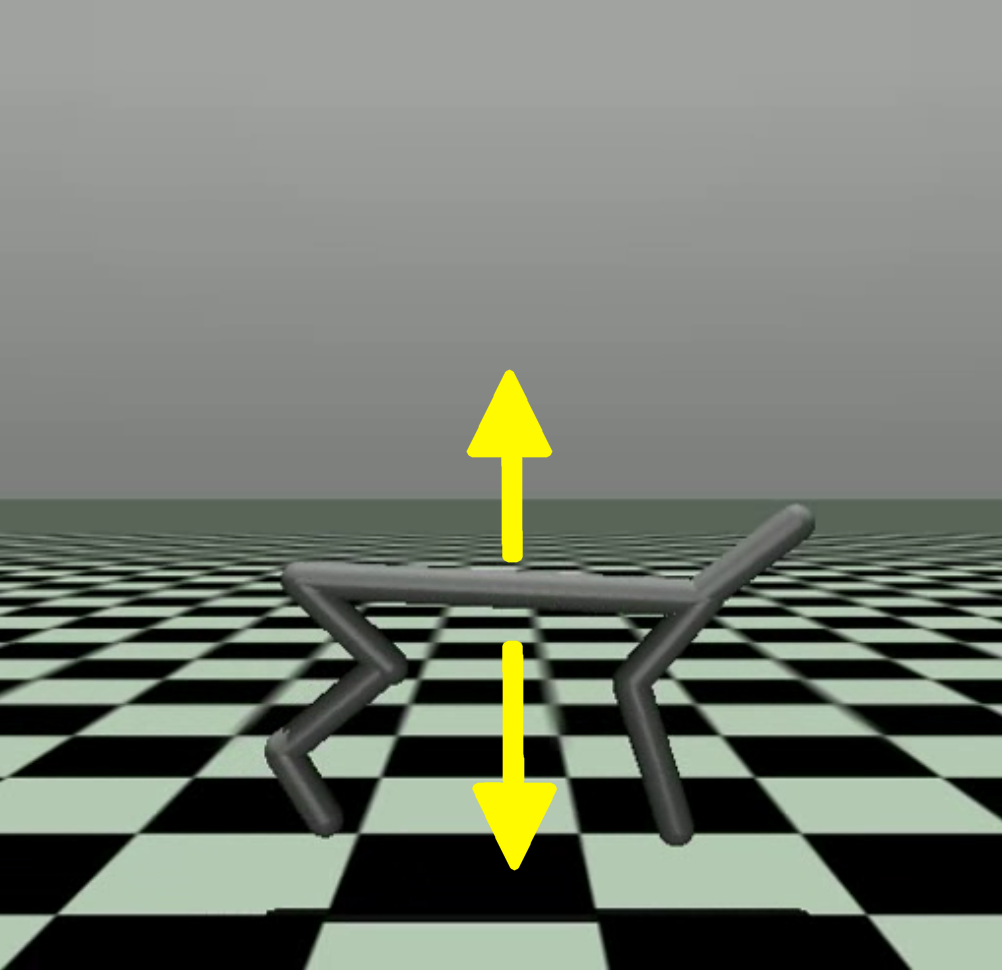}
     }
     \subfloat[][Front flip]{
      \includegraphics[width=0.18\textwidth]{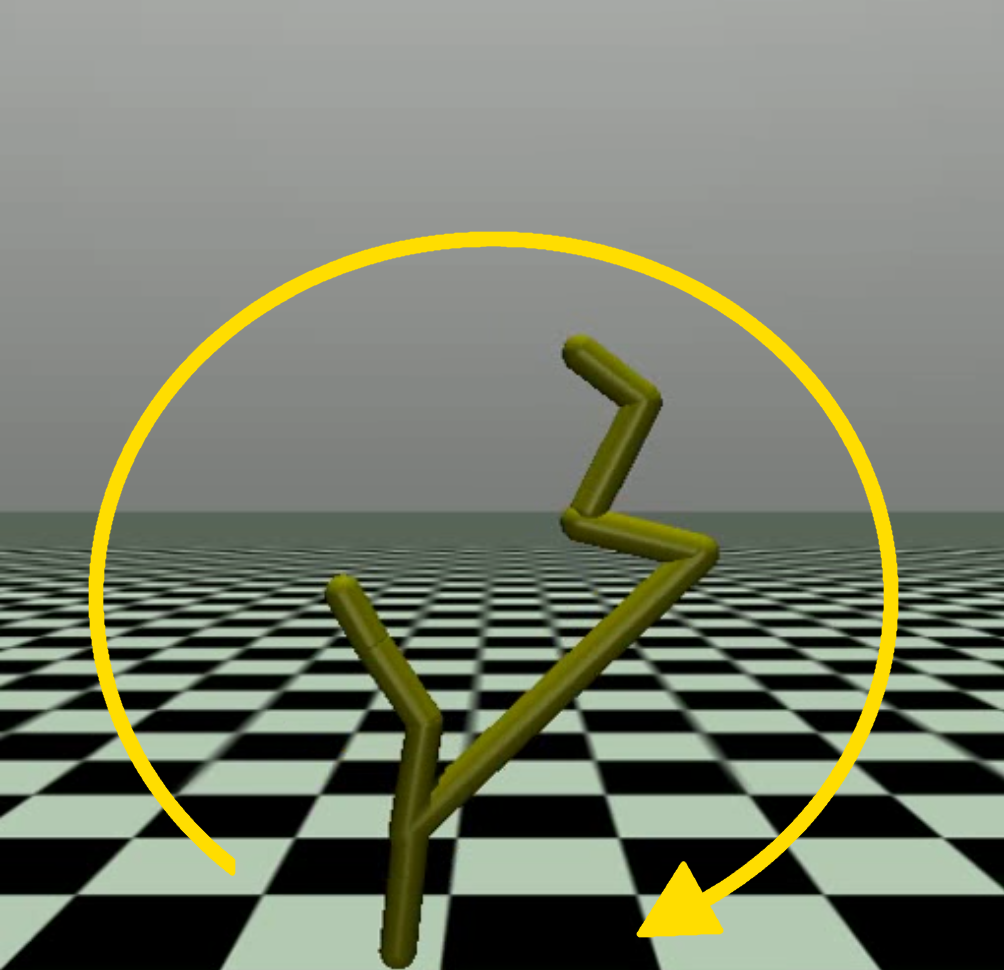}
     }
     \caption{Visualization of eight base environment tasks (yellow arrows show the movement direction).
     These eight tasks share similar dynamic models, but are qualitatively distinct. Each task contains parametric variations, e.g., different goal velocities in the run forward/backward task.}
    \label{fig:environments}
\end{figure*}

Two key properties define the underlying structure of task distributions in meta-RL \cite{metaworld}: parametric and non-parametric variability. Parametric variability describes tasks that qualitatively share the same properties (i.e., their semantic task descriptions are similar), but the parameterization of the tasks varies.
Parametric task distributions tend to be more homogeneous and narrower, which limits the generalization ability of the trained agent to new tasks.
Non-parametric variability, however, describes tasks that are qualitatively distinct but share a common structure, 
so a single meta-RL agent can succeed (See Figure \ref{fig:environments} for a visual example of non-parametric tasks).
Non-parametric task distributions are considerably more challenging, since each distinct task may contain parametric variations \cite{metaworld}.

\subsection{Continual learning}
In continual learning settings, as described in \cite{gmmdavid}, the tasks that an agent has to solve change during the training process. The agent can use information learned from previous, simpler tasks to succeed in new, more complex environments. Retaining learned abilities from earlier time steps is a critical skill for the agent in a continual learning process. Lack of this skill can lead to forgetting of older abilities on the one hand, but can also benefit overfitting on earlier tasks on the other.
We consider a continual learning setting in which access to different non-parametric tasks changes over time, but the agent is given the total number of non-parametric tasks at the start of training. We explore two different possibilities to regulate access to the non-parametric tasks: (1) the agent starts with access to only one non-parametric task, and the number of accessible tasks increases during the training process (''linear'' setting); (2) the agent starts with access to only one non-parametric task, and with each new task, access to the previous task is removed, but the gained experience can still be used (''cut'' setting).

\subsection{Probabilistic embeddings for actor-critic RL (PEARL)}
In task-inference-based meta-RL, the task information that the agent lacks to enable it to behave optimally given a problem $\mathcal{T} \sim p(\mathcal{T})$ is modelled explicitly \cite{metarlasti}.
PEARL \cite{pearl} learns a probabilistic latent variable $\bm z$ that encodes the salient task information given by a fixed-length context variable $\bm c_{1:N}^\mathcal{T}$ containing $N$ recently collected experiences of a task $\mathcal{T}$, which is fed into the policy $\pi_\theta(\bm a \vert \bm s, \bm z)$ trained via soft actor-critic (SAC) \cite{softactorcritic} to solve the presented task.
To encode the task information, an inference network $q_\phi (\bm z \vert \bm c_{1:N}^\mathcal{T})$ is learned with the variational lower bound objective
\begin{equation}
    \mathbb{E}_{\mathcal{T} \sim p(\mathcal{T})} \left[ \mathbb{E}_{\bm z \sim q_\phi(\bm z \vert \bm c_{1:N}^\mathcal{T})} \left[ R(\mathcal{T}, \bm z) + \beta \dkl {q_\phi(\bm z \vert \bm c_{1:N}^\mathcal{T})}{p(\bm z)} \right] \right] \text{,}
\end{equation}
where $p(\bm z)$ is a Gaussian prior used as a bottleneck constraint on the information of $\bm z$, given context $\bm c_{1:N}^\mathcal{T}$ using the KL-divergence $\text{D}_\text{KL}$. 
$R(\mathcal{T}, \bm z)$ is an objective used to train the encoder via the Bellmann critics loss.
$\beta$ is a hyperparameter for weighting the KL-divergence.
The probabilistic encoder in PEARL is modeled as a product of independent Gaussian factors over the $N$ transitions
$
    q_\phi (\bm z \vert \bm c_{1:N}^\mathcal{T}) \propto \prod_{n} \Psi_\phi (\bm z \vert \bm c_n^\mathcal{T})
$,
where 
$
    \Psi_\phi (\bm z \vert \bm c_n^\mathcal{T}) \sim \mathcal{N}(f_\phi^\mu(\bm c_n^\mathcal{T}), f_\phi^{\sigma^2}(\bm c_n^\mathcal{T}))
$
and $f_\phi$ is represented as a neural network with parameters $\phi$ that outputs the mean $\mu$ and variance $\sigma^2$ of the Gaussian conditioned on the context $\bm c_n^\mathcal{T}$. 
In the data collection phase, previous experiences are iteratively added to the context to predict the new latent task representation used in the policy for the next action. 
During testing, PEARL performs a few-shot adaptation by first collecting experiences and then computing a posterior for the latent representation, which remains unchanged during the entire roll-out.
The few-shot mechanism renders PEARL inapplicable to non-stationary task changes.
We highly encourage readers to read about PEARL \cite{pearl} for its intuitive visualizations and in-depth explanations, as preliminaries to this work.







%% file: 03_related_work.tex
\section{Related Work}

The recent work in the domain of meta-RL can be divided into three groups according to the approach taken: gradient-based, recurrence-based, and task-inference-based.

\subsection{Gradient-based} Gradient-based meta-RL approaches such as MAML \cite{maml} and follow-up methods \cite{continuousadaption, Nagabandi2018, modelbasedmeta} are based on finding a set of model parameters during meta-training that can rapidly adapt to achieve large improvements on tasks sampled from a distribution $p(\mathcal{T})$. During the meta-test phase, the initial learned parameters are adjusted to succeed in the task with few gradient steps \cite{maml}.
Gradient-based approaches can be applied to non-stationary environments only if the parameter adaption is performed after every timestep as in \cite{modelbasedmeta}.

\subsection{Recurrence-based} The key element in recurrence-based methods, as in \cite{recurrentmetarl, learningtolearn, asynchronousdrl, rubiks}, is the implementation of a recurrent model that uses previous interactions to implicitly store information that the policy can exploit to perform well on a task distribution. 
By resetting the model at the beginning of each roll-out and recurrently feeding states, actions, and rewards back to the model, the agent can track the interaction history over the entire path in its hidden state and learn how to memorize relevant task information \cite{recurrentmetarl}. 
In meta-training, the model is trained via back-propagation through time. 
In meta-testing, the parameters are fixed, but the agent's internal state adapts to the new task in zero-shot manner \cite{recurrentmetarl}. 

\subsection{Task-inference-based}
In task-inference-based meta-RL, as in PEARL \cite{pearl}, the information that the agent lacks to enable it to behave optimally is modelled explicitly \cite{metarlasti}. MAESN \cite{maesn}, for example, learns a task-dependent latent space that is used to introduce structured noise into the observations to guide the policy's exploration. 
The authors of \cite{tesp} improve on this idea by modelling the task explicitly using an encoder consisting of a gated recurrent unit (GRU) to extract information from a history of transitions, which is given to the policy in addition to the observations. 
In \cite{graphneuralnetwork}, the feature extraction is improved by leveraging a graph neural network that aggregates task information over time and outputs a Gaussian distribution over the latent representation. 
The authors of \cite{gaussdirichlet} further use a combination of a Dirichlet and a Gaussian distribution to model different base tasks with style factors.

%% file: 04_problem_statement.tex
\section{Problem Statement}

\begin{figure*}[t!]
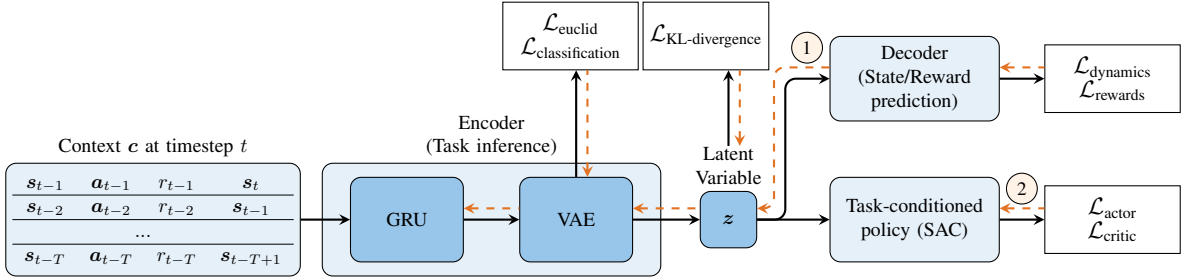

    \drawpipeline
    \caption{Meta-training procedure. The encoder learns a task encoding $\bm z$ from the recent context with gradients from the decoder and provides $\bm z$ for the task-conditioned policy trained via SAC.
    Orange arrows outline the gradient flow.}
    \label{fig:pipeline}
\end{figure*}

This work aims to solve non-parametric meta-RL tasks, in which an agent is trained to maximize the expected discounted return across multiple test tasks from a non-parametric task distribution in a non-stationary setting. 
%
%
%
Specifically, we aim to achieve the following goals:
first, our algorithm should be applicable to non-parametric and broad task distributions in a meta-RL setting. 
Second, the developed algorithm must be able to perform zero-shot adaptation to non-stationary task changes.
Finally, the method should provide the sample efficiency and asymptotic performance of model-free and off-policy algorithms.
To the best of our knowledge, there is no approach that provides the advantages of model-free and off-policy algorithms and is applicable to non-parametric environments in a zero-shot manner.
Currently the only benchmark that satisfies the environment requirements is Metaworld \cite{metaworld}. 
Since Metaworld does not offer well-defined rewards that are normalized across all environments, which can lead to strong bias towards dominant tasks, we provide our own benchmark \textit{half-cheetah-eight} to evaluate different meta-RL approaches.
Clear evidence can be found in the first version of Metaworld \cite{metaworld}, in which very poor performance of state-of-the-art metal RL algorithms \cite{pearl,maml,learningtolearn} are discussed. 

%% file: 05_methodology.tex
\section{Methodology}

In this section, we first give an overview of our
TIGR algorithm. 
We then explain the strategy for making TIGR applicable in non-parametric and non-stationary environments, derive the generative model, and explain how we implement the encoder and decoder. 
Finally, we summarize TIGR with its pseudocode.

\subsection{Overview}
In this paper, we leverage the notion of meta-RL as task inference.
Similar to PEARL, we also extract information from the transition history and use an encoder to generate task embeddings, which are provided to a task-conditioned policy learned via SAC.
We design a generative model for the task inference to succeed in non-parametric environments.
Unlike PEARL, however, we first decouple the training of the probabilistic encoder from the training of the policy by introducing a decoder that reconstructs the underlying MDP of the environment.
Second, we modify the training and testing procedure to encode task information from the recent transition history on a per-time-step basis, enabling zero-shot adaptation to non-stationary task changes.
The structure of our method is shown in Figure \ref{fig:pipeline}.
Our algorithm is briefly explained as follows. 
\begin{itemize}
    \item During meta-training, we first gather interaction experiences from the training tasks and store them in the replay buffer.
    At each interaction, we infer a task representation, such that the policy can behave according to the objective in \eqref{meta_rl_objective}. We feed the recent transition history into a GRU (Sec. \ref{sec:gru_encoder}), which merges the extracted information and forward the features to the VAE (Sec. \ref{sec:gmm}) to generate the overall task representation $\bm z$.
    The task representation is given to the policy with the current observation to predict the corresponding action.
    \item Second, we optimize the task-inference and policy networks, in two sequential stages:
    \begin{itemize}
        \item We first train the GRU-VAE encoder networks for task inference by reconstructing the underlying MDP. For this, we use two additional neural networks that predict the dynamics and reward for each transition (See orange gradient \circled{1} in Figure \ref{fig:pipeline} and Sec. \ref{sec:prediction_model}). This gives the encoder the information required to generate an informative task representation. 
        To improve the performance of the task inference, we employ two additional losses, namely, $\mathcal{L}_\text{classification}$ and $\mathcal{L}_\text{Euclid}$ (See Sec. \ref{sec:clustering_losses}).
        We do not use any gradients from SAC (See orange gradient \circled{2}), which enables us to train the encoder independently of the task-conditioned policy.
        \item In policy training, we compute the task representation for the sampled transitions online using our GRU-VAE encoder. We feed this information to the task-conditioned policy and train it via SAC independently of the task-inference mechanism.
    \end{itemize}
    \item During meta-testing, our method infers the task representation at each timestep, selects actions with the task-conditioned policy and adapts to the task in zero-shot manner.
\end{itemize}

\begin{figure*}[!t]
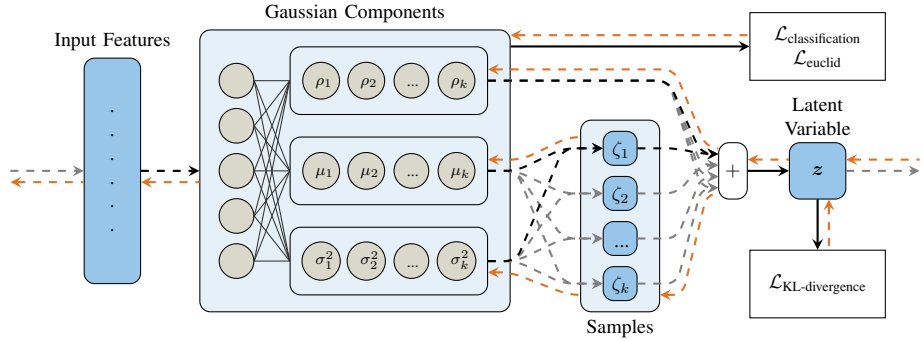

    \drawgmm
    \caption{Overview of the VAE network. The statistics for each Gaussian, including mean $\mu(\bm c, k)$, variance $\sigma^2(\bm c, k)$, and activation $\rho(\bm c, k)$, are computed in parallel. The values are processed and the task representation $\bm z$ is calculated as the weighted sum denoted by the $+$ as described in Sec. \ref{sec:gmm}. 
    Orange arrows outline the gradient flow.
    }
    \label{fig:gmm}
\end{figure*}

\subsection{Task inference}


The non-parametric environments that we consider describe a broad task distribution, with different clusters representing the non-parametric base tasks, and the intra-cluster variance describing the parametric variability for each objective. 
We improve the generative model of PEARL and propose an expressive generative model that captures the multi-modality of the environments and produces reasonable task representations that account for both non-parametric and parametric variability.

\subsubsection{Generative model}\label{sec:generative_model}
Given the sequence of the recent transition history in the last $T$ timesteps as the context $\bm c = (\bm s_{t-T}, \bm a_{t-T}, r_{t-T}, s_{t-T+1} ... , \bm s_{t-1}, \bm a_{t-1}, r_{t-1}, s_t)$, we aim to extract features and find a latent representation $\bm z$ that explains $\bm c \sim p(\bm c \vert \bm z)$ in a generative model such that $p(\bm c, \bm z) = p(\bm c \vert \bm z) p(\bm z)$.
Using the last $T$ timesteps to infer the current task representation $z$ at each interaction allows the model to adapt to task changes online and in zero-shot manner.
We reinterpret the joint structure of the meta-RL task distribution as a combination of different features in a latent space that represent the properties of a particular objective.
Following this idea, we model $\bm z$ as a combination of Gaussians that can express both non-parametric variability with the different Gaussian modes and parametric variability using the variance of a particular Gaussian $k$ with its statistics $\mu(\bm c, k)$ and $\sigma^2(\bm c, k)$. 
We introduce an activation $\rho(\bm c, k)$ that determines the impact of each Gaussian $k$ on $\bm z$ subject to $\sum_k \rho(\bm c, k) = 1$. We sample representatives
\begin{equation}
    \zeta_k \sim \mathcal{N}\left(\mu(\bm c, k), \sigma^2 (\bm c, k)\right)
\end{equation}
from each Gaussian, which are then fused to represent the latent task encoding. We obtain a linear combination of random variables $\zeta_k$ representing different latent task features, which describes a distribution with
\begin{equation}
    \label{eq:gmm}
    p(\bm z\vert \bm c) = \sum_k \rho(\bm c, k)\cdot \zeta_k
\end{equation}
Given a shared structure between tasks, the weights $\rho(\bm c, k)$ thus represent the activation for the latent features that explain the tasks' properties.
The final distribution corresponds to a VAE with a single Gaussian distribution, but we show that the explicit parameterization with multiple Gaussian modes is advantageous for representing non-parametric task distributions. First, linear combination using the activations $\rho(\bm c, k)$ allows us to permanently assign a Gaussian to each task, which is beneficial in continual learning settings because the separate parameters for each Gaussian prevents forgetting of older skills. Second, the explicit parameterization allows us to use prior information about the non-parametric task distribution, since we can train the algorithm to discriminate between non-parametric tasks using a categorical regularizer similar to citegmmdavid for the activations $\rho(\bm c, k)$. We argue that the use of this prior during training does not violate the principle of meta-RL because it is not given directly to the agent.
We compute $\mu(\bm c, k), \sigma^2(\bm c, k)$ and $\rho(\bm c, k)$ as described in the following section.

\paragraph{VAE architecture}\label{sec:gmm}
Using the variational inference approach, we approximate the intractable posterior $p(\bm z \vert \bm c)$ using a variational posterior $q_\theta (\bm z \vert \bm c)$, parameterized by neural networks with parameters $\theta$. The neural network that implements the VAE is designed as a multilayer perceptron (MLP) that predicts the statistics for each Gaussian component including the mean $\mu(\bm c, k)$, variance $\sigma^2(\bm c, k)$ and activation $\rho(\bm c, k)$ as a function of the input features derived from the context $\bm c$ (See Figure \ref{fig:gmm}). 
The standard model is given as a two-layer network and an output layer size of 
$
    K\times (\text{dim}(\bm z) \times 2 + 1)
$,
where $\text{dim}(\bm z)$ is the latent dimensionality required for mean $\mu(\bm c, k)$ and variance $\sigma^2(\bm c, k)$, and the Gaussian activation value is $\rho(\bm c, k)$. $K$ is the number of Gaussian components, which we set equal to the number of non-parametric tasks. This gives the algorithm prior information about the task distribution, which could be circumvented in future work by determining $K$ online as described in \cite{gmmdavid}.
Finally, we represent the VAE components as multivariate Gaussian distributions with mean $\mu(\bm c, k)$ and covariance $\Sigma_k = \bm I \odot \sigma^2(\bm c, k)$, where 
the diagonal of $\Sigma_k$ consists of the entries of $\sigma^2(\bm c, k)$, while every other value is $0$. This assumes that there are no statistical effects between the tasks. We apply a softplus operation to enforce that the network output $\sigma^2(\bm c, k)$ contains only positive values.
We sample from the multivariate Gaussian distributions and obtain representatives $\zeta_k$ for each Gaussian component.
We enforce $\sum_k \rho(\bm c, k) = 1$ by computing the softmax over the VAE's output for the $\rho(\bm c, k)$ values.
Using the computed activations $\rho(\bm c, k)$ and the representatives, we obtain the final latent task representation as in \eqref{eq:gmm}.

\begin{figure}[!t]
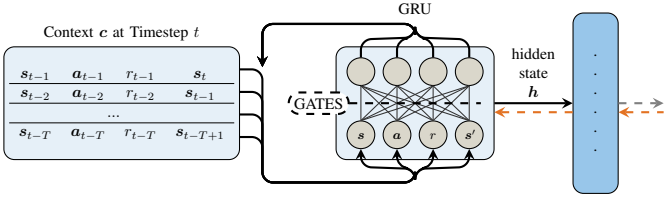

    \centering
    \drawgru
    \caption{Overview of GRU feature extraction. Transitions from the context $\bm c$ are fed in recurrently and the last hidden state is extracted. Orange arrows outline the gradient flow.}
    \label{fig:gru_architecture}
\end{figure}

\paragraph{Feature extraction}\label{sec:gru_encoder}
In this paper, we consider RL environments that are described as high dimensional MDPs. 
To enable our VAE to produce an informative task representation from the high dimensional input data, we first employ a feature extraction mechanism to find the relevant information contained in the context. 
We employ a GRU to process the sequential input data (See Figure \ref{fig:gru_architecture}).
We recurrently feed in the transitions of the context $\bm c$, and thereby combine the features internally in the GRU's hidden state. We extract this hidden state after the last transition is processed and forward it into the VAE.
We implement two more feature-extraction architectures for comparison:
(1)~A shared multilayer perceptron (MLP) architecture similar to PEARL \cite{pearl} that processes each
transition of the context in parallel. The extracted features are passed into the VAE, where we combine the Gaussians for each timestep using the standard Gaussian multiplication with 
$
    \mu = \frac{\mu_1\sigma_2^2 + \mu_2\sigma_1^2}{\sigma_2^2+\sigma_1^2} \text{ and } \sigma^2 = \frac{\sigma_1^2\sigma_2^2}{\sigma_1^2 + \sigma_2^2}
$.
(2)~A transformer architecture \cite{transformer} that creates a key-value embedding for each transition in the context. We extract features from the embedding with a linear layer and forward them to the VAE, where we combine the Gaussians for each timestep using the standard Gaussian multiplication (See MLP).
We provide an ablation study of the feature-extraction configurations in the discussion section.

\subsubsection{Encoder-decoder strategy}
\label{sec_encoder_decoder}
Our generative model follows the idea of a VAE. 
The setup employs an encoder, to describe the latent task information given a history of transitions from an MDP as $q_\theta(\bm z \vert \bm c)$; and a decoder to reconstruct the MDP from the latent task information given by the encoder as $p_\phi(\bm c \vert \bm z)$. 
The encoder is modeled as a VAE which involves the prior use of a shared feature extraction method, as described in the previous section. The decoder implements the generating function $p_\phi(\bm c \vert \bm z)$, parameterized as neural networks with parameters $\phi$. 
Following the variational approach in \cite{vae}, we derive the evidence lower bound objective (ELBO) for the encoder and decoder 
and obtain:
\begin{align}
    \log p_\phi(\bm c) &\geq \mathcal{L}(\theta,\phi;\bm c)  \nonumber\\
    &= \mathbb{E}_{q_\theta (\bm z \vert \bm c)} \left[ \log p_\phi(\bm c, \bm z) - \log q_\theta (\bm z \vert \bm c) \right] \nonumber\\
    &= \mathbb{E}_{q_\theta (\bm z \vert \bm c)}\left[ \log p_\phi(\bm c \vert \bm z) \right] - \dkl{q_\theta (\bm z \vert \bm c)}{p_\phi (\bm z)} \text{.}
\end{align}
We use the reparameterization trick and combine the $k$ Gaussian components to arrive at
\begin{equation}
    \label{eq:reparametrize}
    \Tilde{\bm z} = \sum_k \rho_{q_\theta}(\bm c, k) \left(\mu_{q_\theta}(\bm c, k) + \epsilon\cdot\sigma^2_{q_\theta}(\bm c, k) \right)
\end{equation}
with $\epsilon \sim \mathcal{N}(0, 1)$ and apply Monte Carlo sampling to arrive at the objective:
\begin{equation}
    \label{eq:elbo}
    \mathcal{L}(\theta, \phi; \bm c) \approx \bigg[ \log p_\phi (\bm c \vert \Tilde{\bm z}) - \dkl{q_\theta (\Tilde{\bm z} \vert \bm c)}{p_\phi(\Tilde{\bm z})} \bigg] \text{,}
\end{equation}
where $\log p_\phi (\bm c \vert \Tilde{\bm z})$ is a reconstruction objective of the input data $\bm c$.
$\dkl{q_\theta (\Tilde{\bm z} \vert \bm c)}{p_\phi(\Tilde{\bm z})}$ introduces a regularization conditioned on the prior $p_\phi(\Tilde{\bm z})$.
Since the lower bound objective requires maximization, we denote the optimization objective as minimizing $-\mathcal{L}(\theta, \phi; \bm c)$, which results in a negative log-likelihood objective for the reconstruction term.

\paragraph{Reconstruction objective}\label{sec:prediction_model}
The reconstruction objective provides the information necessary for the encoder and VAE to extract and compress relevant task information from the context. 
It can take many forms, as suggested in \cite{pearl}, such as reducing the Bellmann critic's loss, maximizing the actor's returns, and reconstructing states and rewards.
We follow the third proposal and extend the negative log-likelihood objective of reconstructing states and rewards to predicting the environment dynamics and the reward function for the underlying MDP (See Appendix Figure \ref{fig:prediction_architecture}). 
We split the decoder into two parts $p_{\phi_\text{dynamics}}$ and $p_{\phi_\text{rewards}}$ modeled as MLPs (See Figure \ref{fig:prediction_architecture}), which predict the next state $\bm s'$ and reward $r$ given $\bm s, \bm a$ and $\bm z$, and train them with the following loss:
\begin{align}
    \log p_\phi(\bm c \vert \Tilde{\bm z}) &= \log p_\phi(\bm s',r \vert \bm s, \bm a, \Tilde{\bm z}) \nonumber\\
    &= \log p_{\phi_\text{dynamics}}(\bm s' \vert \bm s, \bm a, \Tilde{\bm z}) + \log p_{\phi_\text{rewards}}(r \vert \bm s, \bm a, \Tilde{\bm z}) \text{.}
\end{align}
We model both parts as regression networks, in which the data is modeled as a normal distribution. Thus, the loss function is defined as the sum of $\mathcal{L}_\text{dynamics}(\phi)$ and $\mathcal{L}_\text{rewards}(\phi)$ as:
\begin{align}
    \mathcal{L}_{\text{prediction}}(\phi)
    = &\frac{1}{\text{dim}(\bm s)} \norm{\bm s' - p_{\phi_\text{dynamics}}(\bm s' \vert \bm s, \bm a, \Tilde{\bm z})}^2 \nonumber\\
    + &\frac{1}{\text{dim}(r)} \norm{r - p_{\phi_\text{rewards}}(r \vert \bm s, \bm a, \Tilde{\bm z})}^2
\end{align}
where both components are normalized by the number of their dimensions.

\begin{figure}[!t]
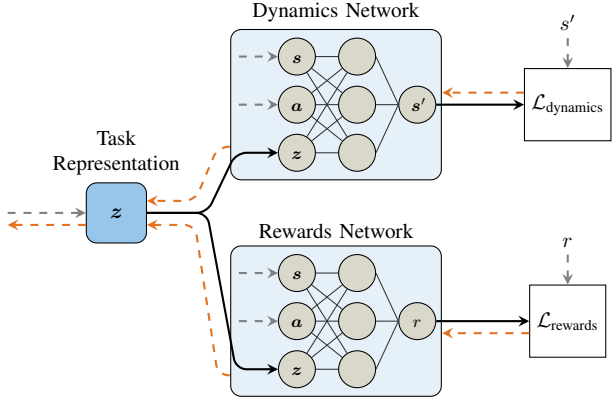

    \centering
    \drawprediction
    \caption{Overview of dynamics and reward prediction networks. The latent task representation $\bm z$ in addition to the state $\bm s$ and action $\bm a$ are used as the input to both networks. Orange arrows outline the gradient flow.}
    \label{fig:prediction_architecture}
\end{figure}

\paragraph{Information bottleneck}
The regularization introduced by the KL-divergence in \eqref{eq:elbo} serves as an information bottleneck that helps the VAE compress the input to a compact format.
Due to the potential over-regularization of this term \cite{gmmvae}, we control its impact on the ELBO as
\begin{equation}
    -\mathcal{L}(\theta, \phi;\bm c) \approx \mathcal{L}_\text{NLL}(\theta, \phi) + \alpha\cdot \mathcal{L}_\text{KL-divergence}(\theta)
\end{equation}
with a factor $\alpha < 1$ to allow expressive latent representations.

\paragraph{Clustering losses} \label{sec:clustering_losses}

To improve the task inference performance for non-parametric environments, we employ two additional losses that represent the following ideas.

\begin{itemize}
    \item Assign each component of the VAE to one base task only. This objective is related to component-constraint learning as in \cite{gmmdavid}. The idea is to enforce each Gaussian to represent one base task, containing only task-specific features. We can feed prior information about a true base task $y$ to the algorithm and use a supervised classification learning approach for the activations $\rho(\bm c, k)$ with the standard cross-entropy formulation as
    \begin{equation}
        \mathcal{L}_\text{classification}(\theta) = -\log \left( \frac{\exp(\rho(\bm c, k = y))}{\sum_k \exp(\rho(\bm c, k))} \right)
    \end{equation}
    Cross-entropy enforces that task activations correspond to the base task distribution such that $\rho_k \xrightarrow[]{} 1$ for $k \overset{\Delta}{=} y_k$. By introducing this secondary objective, we have the possibility to constrain the components to represent task-specific features instead of shared features among tasks.
    \item Push the components of the VAE away from each other to achieve a clear distinction between base tasks.
    To be able to further distinguish the features and prevent overlap, we confine them to separate clusters. This can be realized by an objective that seeks to maximize the euclidean distance between the means $\mu(\bm c, k)$ of the $K$ components scaled by the sum of variances $\sigma^2(\bm c, k)$ with
    \begin{equation}
        \mathcal{L}_\text{Euclid}(\theta) = \sum_{k_1=1}^K \sum_{k_2=k_1+1}^K \frac{\sigma^2(\bm c, k_1) + \sigma^2(\bm c, k_2)}{\norm{\mu(\bm c, k_1) - \mu(\bm c, k_2)}^2}
    \end{equation}
    The Euclidean distance is replaced by the sum of squares, which avoids the computationally unstable calculation of the square root, but has the same effect.
\end{itemize}

We provide an evaluation of the impact of the clustering losses in the experiments section.

\begin{algorithm}[!t]
\caption{TIGR Meta-training}\label{alg:ours}
\begin{algorithmic}[1]
    \Require Encoder $q_\theta$, decoder $p_\phi$, policy $\pi_\psi$, Q-network $Q_\omega$, task distribution $p(\mathcal{T})$, replay buffer $\mathcal{D}$
    \For{each epoch}
        \State Perform roll-out for each task $\mathcal{T} \sim p(\mathcal{T})$, store in $\mathcal{D}$
        \For{each task-inference training step}\label{algln:ti_start}
            \State Sample context $\bm c\sim\mathcal{D}$
            \State Compute $\bm z = q_\theta(\bm c)$ \Comment{(See Sec. \ref{sec:gmm})}
            \State Calculate losses $\mathcal{L}_\text{KL-divergence}$, $\mathcal{L}_\text{Euclid}$, $\mathcal{L}_\text{classification}$ \Comment{(See Sec. \ref{sec:finalobjective})}
            \State Compute $(\bm s', r) = p_\phi(\bm s,\bm a,\bm z)$ \Comment{(See Sec. \ref{sec:prediction_model})}
            \State Calculate loss $\mathcal{L}_\text{prediction}$
            \State Derive gradients for losses with respect to $q_\theta,p_\phi$ and perform optimization step
        \EndFor\label{algln:ti_end}
        \For{each policy training step}\label{algln:policy_start}
            \State Sample RL batch $d\sim\mathcal{D}$ and corresponding context $\bm c$, infer $\bm z = q_\theta(\bm c)$
            \State Perform SAC algorithm for $\pi_\psi, Q_\omega$ 
        \EndFor\label{algln:policy_end}
    \EndFor
    \Statex
    \Return $q_\theta, \pi_\psi$
\end{algorithmic}
\end{algorithm}

\paragraph{Final objective}
\label{sec:finalobjective}
The final objective derived from the reconstruction objective, information bottleneck, and clustering losses is used to jointly train the encoder-decoder setup as described in Section \ref{sec_encoder_decoder}. We combine the different loss functions to enable the encoder to produce informative embeddings, which are used in the task-conditioned policy. The resulting overall loss is denoted as:
\begin{align}
    \mathcal{L}(\theta, \phi) = &\mathcal{L}_\text{NLL (prediction)}(\theta, \phi) + \alpha\cdot \mathcal{L}_\text{KL-divergence}(\theta)\nonumber\\ 
    &+ \beta\cdot \mathcal{L}_\text{Euclid}(\theta) + \gamma\cdot \mathcal{L}_\text{classification}(\theta)
\end{align}
where $\alpha, \beta, \gamma$ are hyper-parameters that weigh the importance of each term.

\subsection{Algorithm overview}

The TIGR algorithm is summarized in pseudo-code (See Algorithm \ref{alg:ours}).
The task-inference mechanism is implemented from line~\ref{algln:ti_start} to line~\ref{algln:ti_end}. 
Lines~\ref{algln:policy_start} and~\ref{algln:policy_end} implement the standard SAC~\cite{softactorcritic}. A list of the most important hyperparameters of the algorithm and their values is given in Appendix \ref{appendix_evaluation}.

%% file: 06_experiments.tex
\section{Experiments}
\label{chapter:experiments}
%

%
%

We evaluate the performance of our method on the non-parametric \textit{half-cheetah-eight} benchmark that we provide and verify its wide applicability on a series of other environments.
The \textit{half-cheetah-eight} environments are shown in Figure \ref{fig:environments} and their detailed descriptions are introduced in Appendix \ref{appendix_env}.
The evaluation metric is the average reward during the meta testing phase. 
We first compare the sample efficiency and asymptotic performance against state-of-the-art meta-RL algorithms, including PEARL. 
Second, we visualize the latent space encoding of the VAE. 
Third, we evaluate the task-inference capabilities of our algorithm in the zero-shot setup.
Fourth, we evaluate the applicability of the algorithm to non-stationary task changes in the \textit{half-cheetah-eight} benchmark.
Finally, we provide videos displaying the distinct learned behaviors in the supplementary material, along with our code.

\begin{figure*}[!t]
    \centering
    \includegraphics[width=0.95\textwidth]{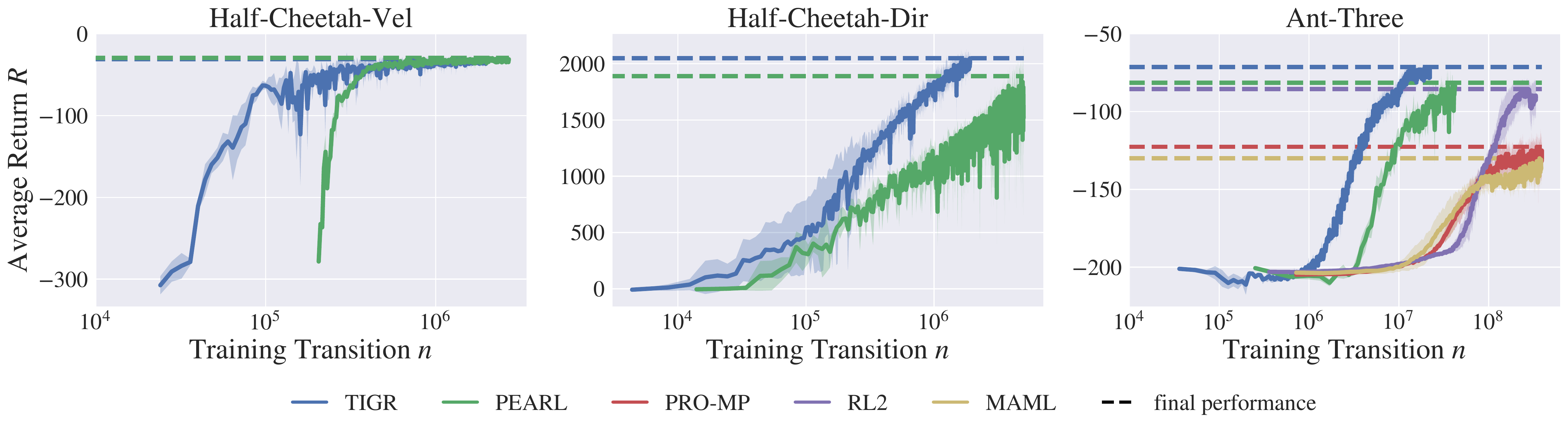}
    \caption{Meta-testing performance over environment interactions evaluated periodically during the meta-training phase. As PEARL \cite{pearl} outperforms ProMP \cite{promp}, RL2 \cite{learningtolearn}, and MAML \cite{maml} in \textit{half-cheetah-vel} and \textit{half-cheetah-dir}, we only compare TIGR with PEARL in these two environments.
    The blue line shows the performance of our method. We show the mean performance over three independent runs, evaluated in the \textit{half-cheetah-velocity}, \textit{half-cheetah-direction}, and \textit{ant-three} environments. Note that the x-axis is in \textbf{log scale}.
    }
    \label{fig:single_performance}
\end{figure*}
\begin{figure*}[!t]
    \centering
    \subfloat[][Meta-testing performance]{
       \includegraphics[width=0.45\textwidth]{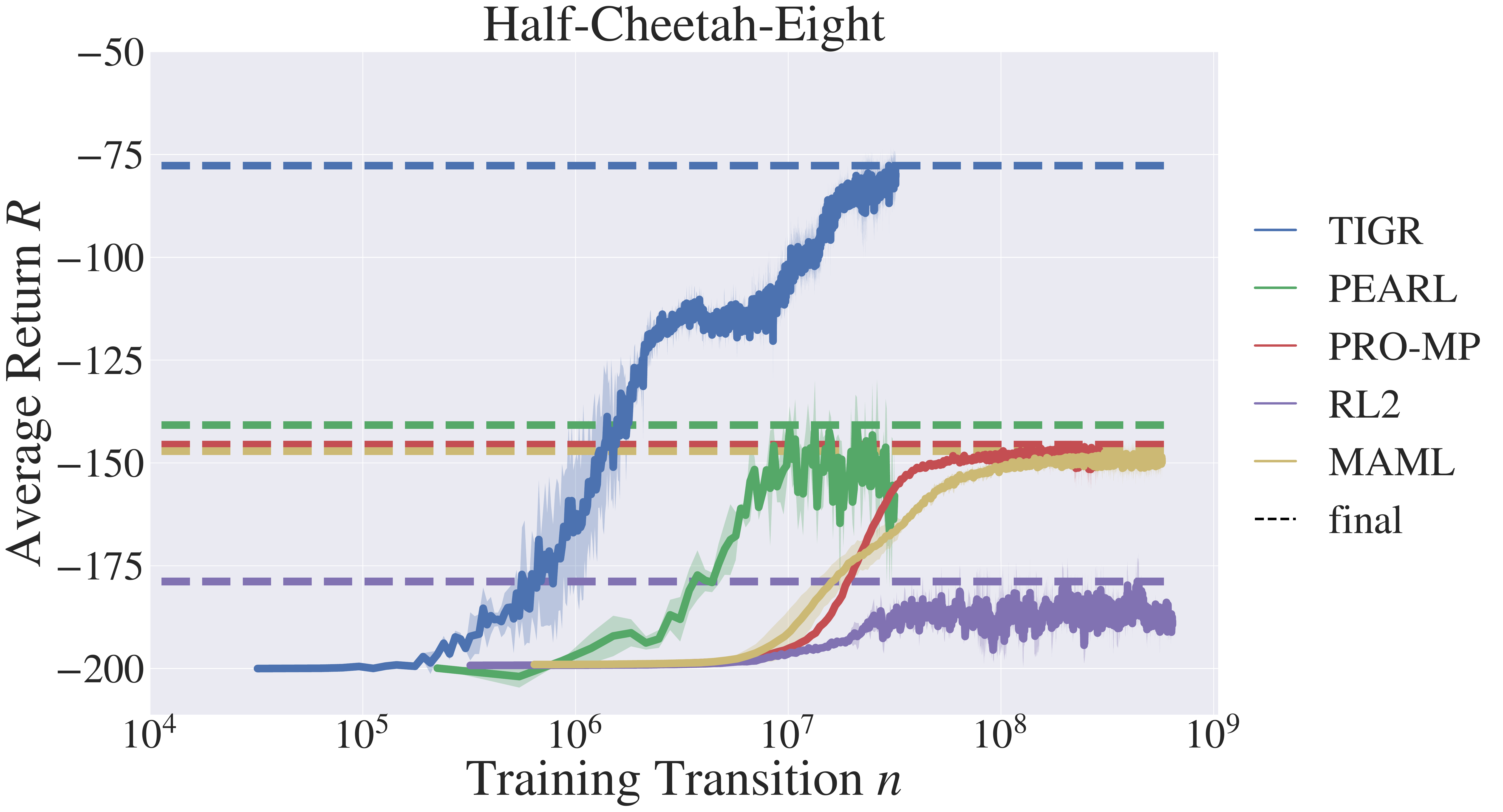}
    \label{fig:test_performance}
    }
    \subfloat[][Latent encoding]{
       \includegraphics[width=0.45\textwidth]{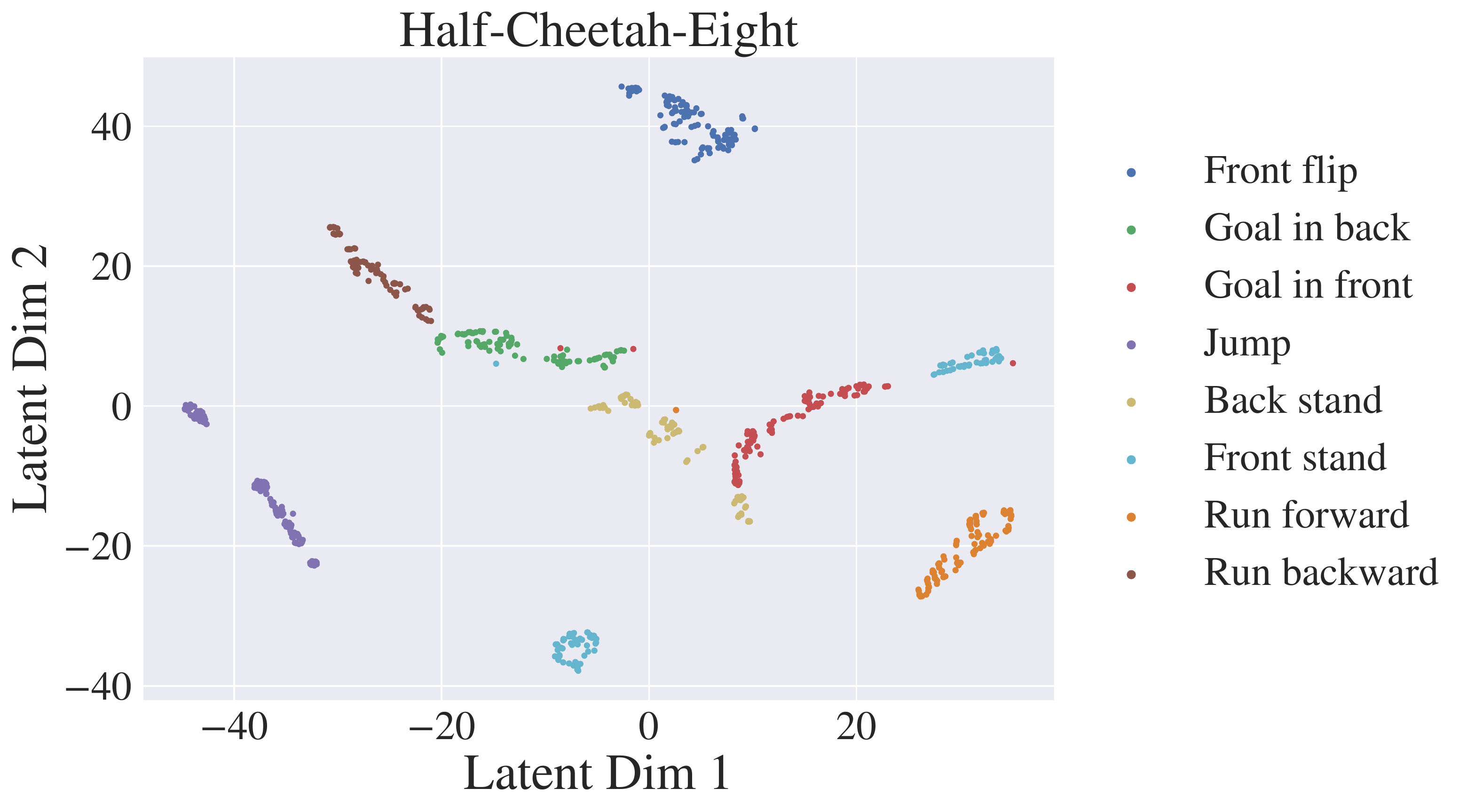}
    \label{fig:latent_encoding}
    }
    \caption{(a) Meta-testing performance over environment interactions evaluated periodically during the meta-training phase. 
    We show the mean performance from three independent runs.
    (b) Final encoding of the eight tasks visualized using T-SNE \cite{van2008visualizing} in two dimensions.
    %
    Note that the x-axis is in \textbf{log scale}.
    }
\end{figure*}

\subsection{Asymptotic performance and sample efficiency} 
We first demonstrate the performance of PEARL and our method in standard parametric environments, namely \textit{half-cheetah-vel} and \textit{half-cheetah-dir} tasks \cite{pearl}, to verify both approaches.
It should be noted that PEARL achieves reported performances in few-shot manner while our method is tested at first sight in zero-shot fashion.
Figure \ref{fig:single_performance} shows that both methods achieve similar performance and can solve the tasks. However, TIGR significantly outperforms PEARL in terms of sample efficiency across both tasks, even in zero-shot manner.

We then progress to a slightly broader task distribution and evaluate the performance of PEARL, other state-of-the-art algorithms, and our method, on a modification of the \textit{ant} environment (a detailed description is introduced in Appendix \ref{appendix_env}). We use three different tasks, namely goal tasks, velocity tasks and a jumping task. The goal and velocity tasks are each split into left, right, up, and down and include different parametrizations.
We take the original code and parameters provided from PEARL \cite{pearl}. For a fair comparison, we adjust the dimensionality of the latent variable to be the same. For the other meta-RL algorithms, we use the code provided by the authors of Pro-MP~\cite{promp}.\footnote{Repository available at \url{https://github.com/jonasrothfuss/ProMP/tree/full_code}.}
Figure \ref{fig:single_performance} on the right shows the performances of these five methods. We see that although PEARL and RL2 can solve the different tasks, our method greatly outperforms the others in sample efficiency and has a slight advantage in asymptotic performance.

Finally, we evaluate the approaches on the \textit{half-cheetah-eight} benchmark.
The average reward during meta-testing is shown in Figure \ref{fig:test_performance}. 
We can see that TIGR outperforms prior methods in terms of sample efficiency and demonstrates superior asymptotic performance. 
Looking at the behaviours showcased in the video provided with the supplementary material, we find that with a final average return of $-150$, the PEARL agent is not able to distinguish the tasks. 
We observe a goal-directed behavior of the agent for the goal tasks but no generalization of the forward/backward movement to velocity tasks. 
It learns how to stand in the front but fails in \textit{stand back}, \textit{jump} and \textit{front flip}.
For TIGR, we can see that every base task except the \textit{front flip} is learnt correctly.
It should be noted that for the customized flip task, we expect the agent to flip at different angular velocities, which is much harder than the standard flip task, in which the rotation speed is simply maximized.

\subsection{Clustering losses}\label{sec:clustering}
\begin{figure}[!t]
    \centering
    \includegraphics[width=0.45\textwidth]{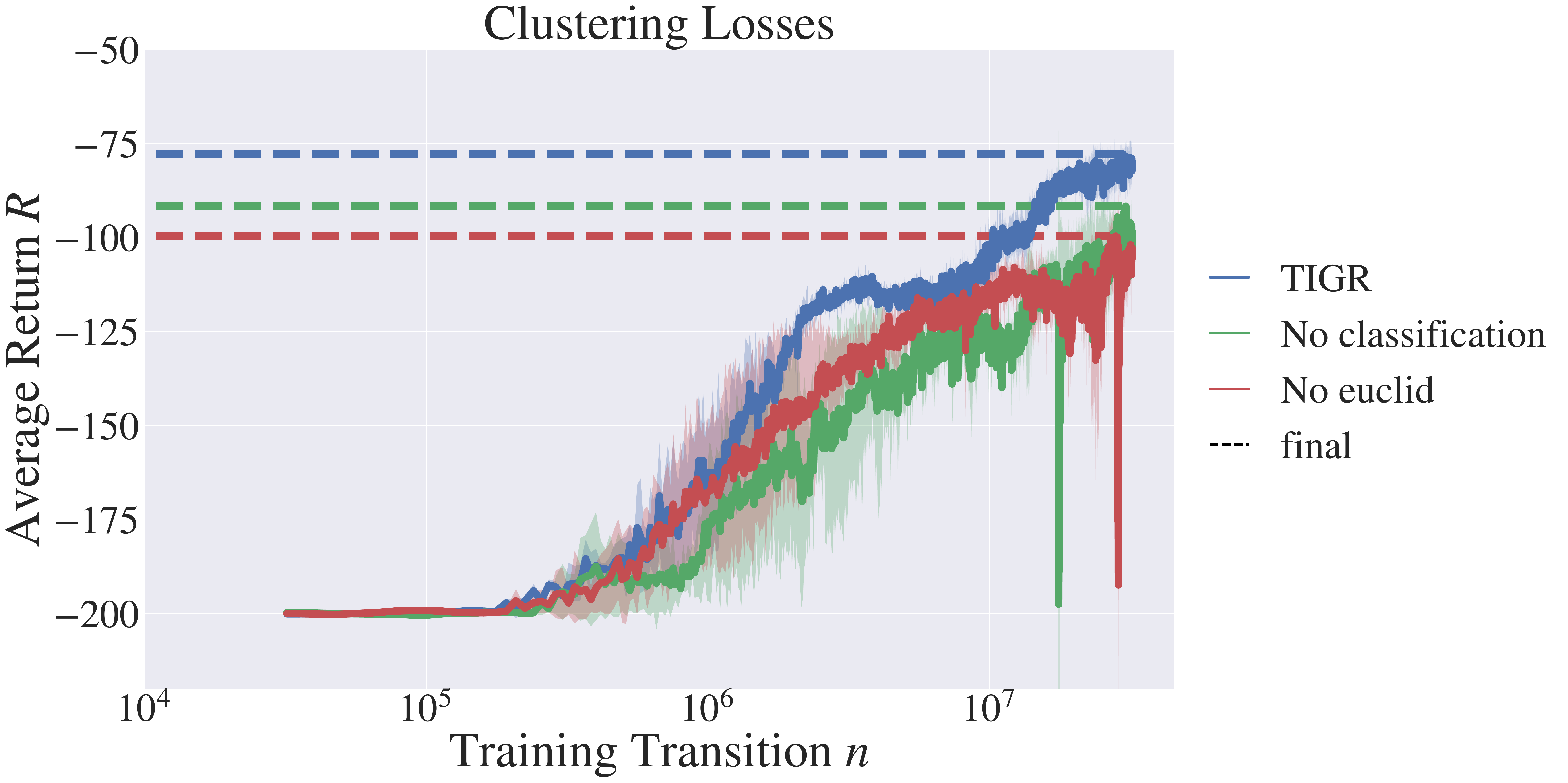}
    \caption{Evaluation of the impact of the proposed clustering losses on the meta-testing performance of the algorithm on the \textit{half-cheetah-eight} benchmark. We remove each of the losses in turn and compare to the setup with all losses involved.}
    \label{fig:clustering_losses}
\end{figure}

We evaluate the impact of the clustering losses on the meta-testing performance of TIGR on the \textit{half-cheetah-eight} benchmark in Figure \ref{fig:clustering_losses}. We see that, when leaving out any one of the losses, the performance is weaker and less stable when one of the losses is omitted. Nevertheless, the algorithm significantly outperforms the prior meta-RL methods in both cases. Thus, each loss has a beneficial impact on the meta-RL objective, but the additional prior information about the true base tasks during meta-training is not mandatory for the algorithm to succeed in the non-parametric environment.

\subsection{Latent space encoding} We evaluate the latent task representation by sampling transition histories from the replay buffer that belong to individual roll-outs. We extract features from the context using the GRU encoder and obtain the compressed representation from the VAE.
The latent task encoding is visualized in Figure \ref{fig:latent_encoding}. 
We use T-SNE \cite{van2008visualizing} to visualize the eight-dimensional encoding in two dimensions. 
The representations are centered around $0$, which demonstrates the information bottleneck imposed by the KL-divergence.
We can see that qualitatively different tasks are clustered into different regions (e.g., \textit{run forward}), verifying that the VAE is able to separate different base tasks from each other.
Some base tasks show clear directions along which the representations are spread (e.g., \textit{run backward}), suggesting how the parametric variations in each base task are encoded.

\subsection{Task inference}
\label{sec:task_inference} 
\begin{figure*}[!t]
    \centering
    \includegraphics[width=\textwidth]{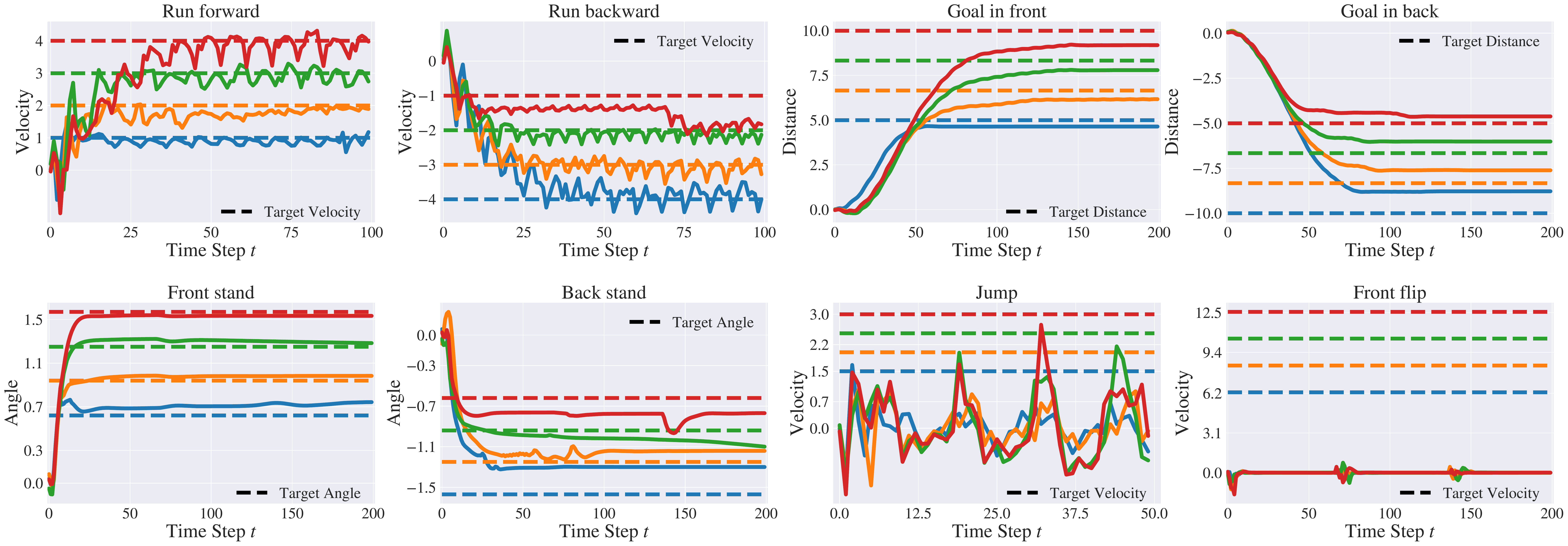}
    \caption{Task-inference response during one episode for the \textit{half-cheetah-eight} benchmark after 2000 training epochs. Each task is evaluated under different parametric variations. The target is marked with a the dashed line. The task is inferred correctly when the solid line approaches the target. The vertical velocity for the \textit{jump} task oscillates due to gravity. The \textit{front flip} task remains unsolved.}
    \label{fig:task_inference_complete}
\end{figure*}

We evaluate whether the algorithm infers the correct task by examining the evolution of the velocity, distance, or angle during an episode. A task is correctly inferred when the current value approaches the target specification.
We can see in Figure \ref{fig:task_inference_complete} that the target specification is reached within different time spans for the distinct tasks. 
This is because goal-based tasks take longer to perform. 
Nevertheless, we can see that the task inference is successful, as the current value approaches the target specification in the displayed settings. For \textit{jump}, the vertical velocity oscillates due to gravity and cannot be kept steady at the target. The \textit{front flip} task remains unsolved.

\begin{figure*}[!t]
    \centering
    \includegraphics[width=\textwidth]{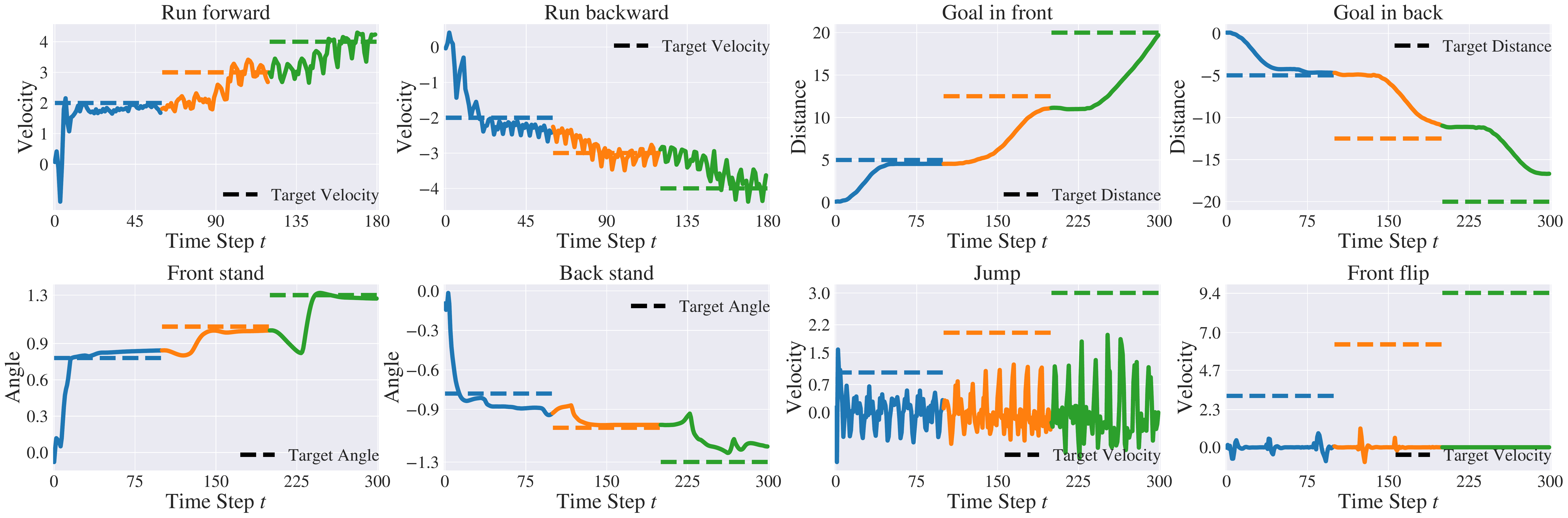}
    \caption{Evaluation of the performance of the TIGR algorithm for parametric non-stationary task changes for each base task in the \textit{half-cheetah-eight} benchmark. The targets (specifications) for each of the environments in the sub-figures are marked as the dashed line. The non-stationary adaptation to the tasks is successful when the solid line for the current value repeatedly approaches the targets. The vertical velocity for the \textit{jump} task oscillates due to gravity. The \textit{front flip} task remains unsolved.}
    \label{fig:single_non_stationary}
\end{figure*}
\begin{figure*}[!t]
    \centering
    \includegraphics[width=\textwidth]{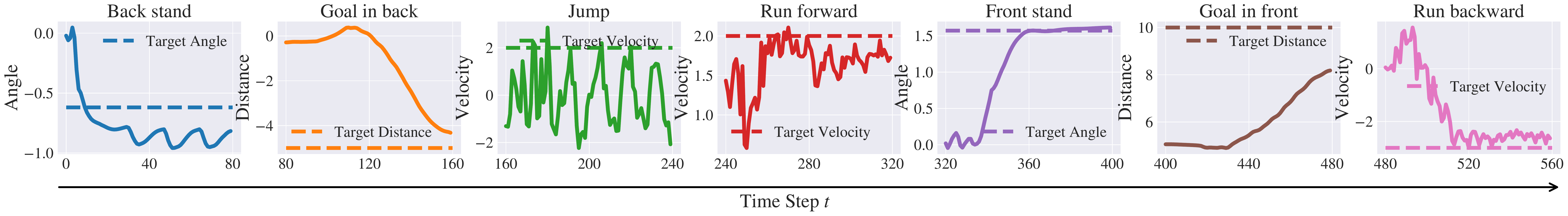}
    \caption{Evaluation of the performance of the TIGR algorithm for non-stationary and non-parametric task changes in the \textit{half-cheetah-eight} benchmark. The target (specification) for each of the environments in the sub-figures is marked as the dashed line. The non-stationary adaptation to the tasks is successful when the solid line for the current value repeatedly approaches the target.}
    \label{fig:non_stationary}
\end{figure*}

\subsection{Applicability to non-stationary environments}
The zero-shot adaptation mechanism of the TIGR algorithm allows us to evaluate its performance for non-stationary task changes in the \textit{half-cheetah-eight} benchmark, which is not possible for few-shot methods as PEARL \cite{pearl}.
First, we evaluate the adaptation to parametric task changes for the eight base tasks in the \textit{half-cheetah-eight} benchmark. We use three consecutive parameterizations each, without resetting the state of the environment or agent in-between. We evaluate whether the algorithm infers the correct tasks by examining the evolution of the velocity, goal distance or angle during the episode. The non-stationary adaptation to the tasks is successful when the current value repeatedly approaches the target specification.
The results are shown in Figure \ref{fig:single_non_stationary}. Each sub-figure depicts a different base task. We find that non-stationary adaptation to the different parameterizations is successful for each base task, as the inspected value repeatedly approaches the target specification the displayed settings, exhibiting similar task-inference patterns as described in section \ref{sec:task_inference}. The vertical velocity for the \textit{jump} task oscillates due to gravity. The \textit{front flip} task remains unsolved.

Second, we evaluate the adaptation to non-parametric task changes in the \textit{half-cheetah-eight} benchmark. We set a fixed order for the base tasks as visualized in Figure \ref{fig:non_stationary} and iterate through them online after executing 80 steps for each environment, without resetting its state in-between. We evaluate whether the algorithm infers the correct tasks by examining the evolution of the velocity, goal distance or angle during the entire episode. The non-stationary adaptation to the tasks is successful when the current value repeatedly approaches the target specification.
The results are shown in Figure \ref{fig:non_stationary}. Each sub-figure describes the evolution of the values for the given base task and its specified target parameterization with successive timesteps across all environments. We find that non-stationary adaptation to the tasks is successful, as the inspected value repeatedly approaches the target specification in all of the displayed settings.

\subsection{Applicability to continual learning}
\begin{figure*}[!t]
    \centering
    \includegraphics[width=\textwidth]{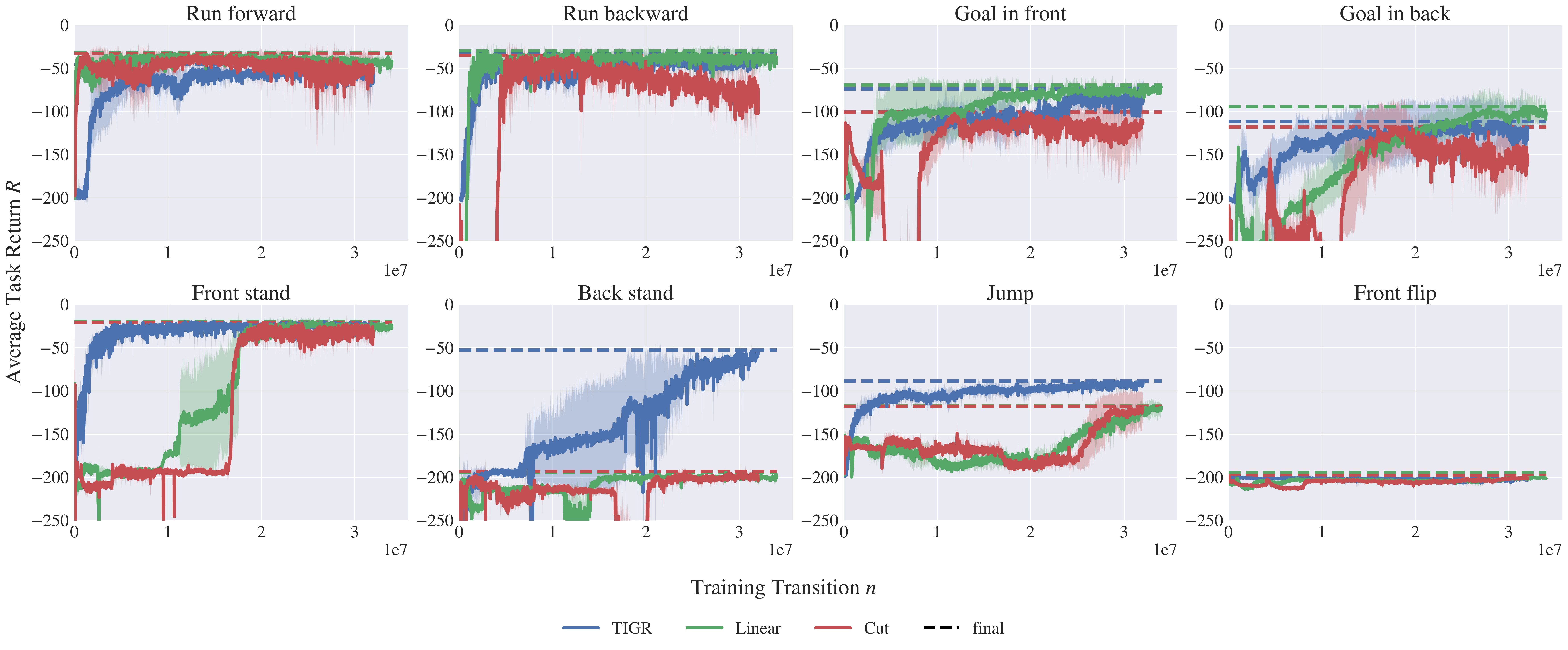}
    \caption{Evaluation of the performance of the TIGR algorithm for the two continual learning settings in the \textit{half-cheetah-eight} benchmark. The non-continual learning curve for TIGR is shown for comparison in blue. The continual learning is successful when the curve for the ''linear'' (green) or ''cut'' (red) setting approaches the TIGR curve.}
    \label{fig:continual_learning}
\end{figure*}

We evaluate the applicability of the TIGR algorithm to two different continual learning settings derived from the \textit{half-cheetah-eight} benchmark: (1) the agent starts with access to only one non-parametric task, and the number of accessible tasks increases during the training process (''linear'' setting); (2) the agent starts with access to only one non-parametric task, and with each new task, access to the previous task is removed, but the experience gained with the previous tasks can still be used (''cut'' setting). The total amount of environment interactions gathered in training is the same for all the displayed settings. We set the number of Gaussians in the VAE equal to the total number of non-parametric tasks presented during the learning process. We use the following order of tasks: \textit{Run forward}, \textit{Run backward}, \textit{Reach front goal}, \textit{Reach back goal}, \textit{Front stand}, \textit{Back stand}, \textit{Jump}, and \textit{Front flip}.
The results are shown in Figure \ref{fig:continual_learning}. Each sub-figure represents a different base task. After each $12.5\%$ of training progress, a new base task is provided in the given order. We note that the agent does not perform as well in the continual learning setting as in the non-continual setting because the reduced training time for difficult tasks that are not accessed until the later stages of training, such as the \textit{Back stand} or \textit{Jump} tasks, does not give the algorithm enough time to fully learn the desired behavior. Nevertheless, the algorithm does not exhibit forgetting or overfitting on old tasks, thus fulfilling these critical abilities necessary for continual learning.

\section{Discussion and Ablation Study}

We perform an ablation study of different configurations of our method on the environments that we provide.
We first compare the sample efficiency and asymptotic performance for the different clustering losses.
Second, we evaluate the performances of the three implemented feature-extraction configurations.
Finally, we discuss the performance and limitations of the TIGR algorithm.

\subsection{Feature extraction ablation}

\begin{figure}[!t]
    \centering
    \includegraphics[width=0.45\textwidth]{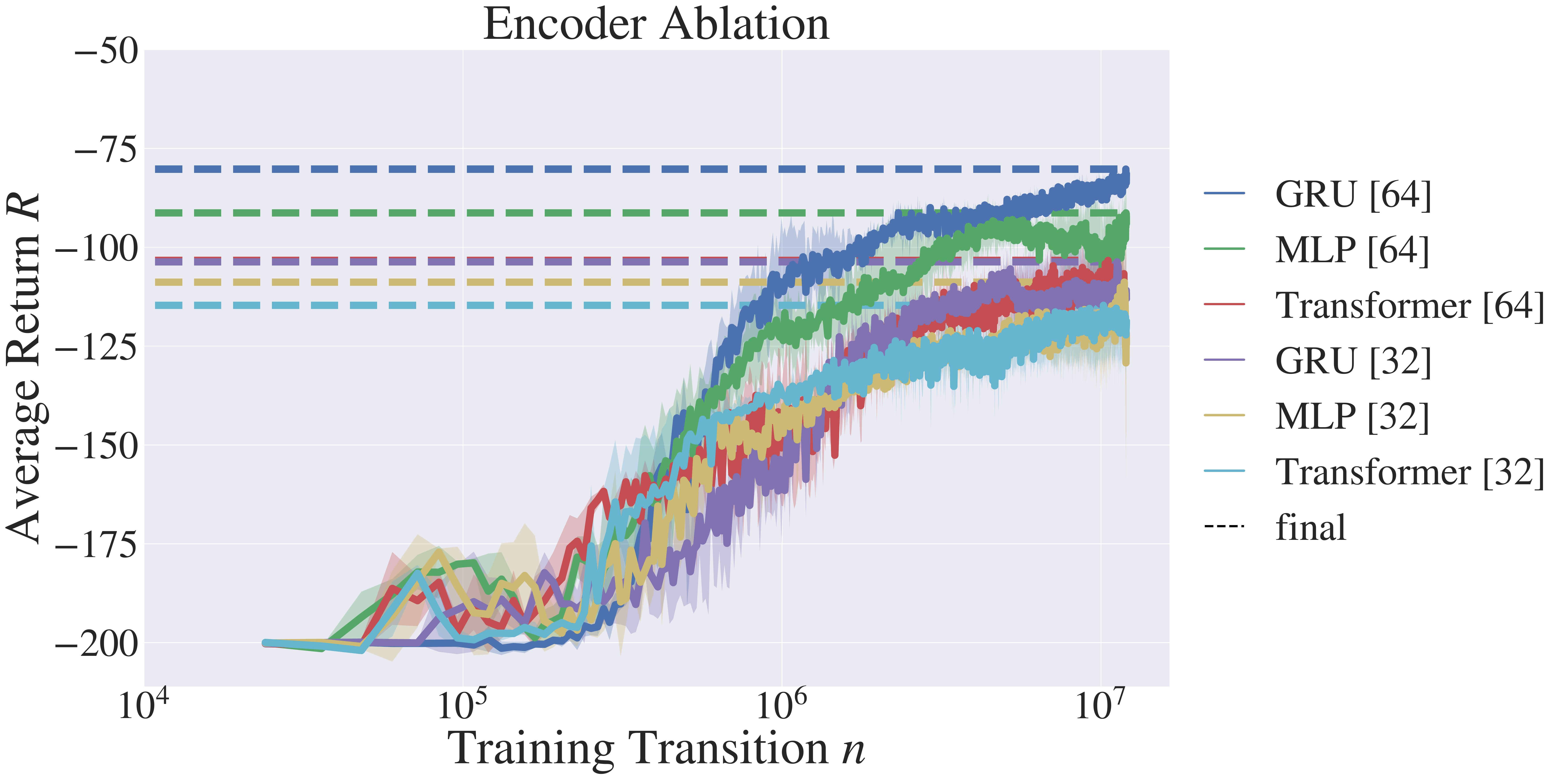}
    \caption{Evaluation of meta-testing performance on the \textit{half-cheetah-six environment} of different feature-extraction configurations in addition to different context lengths in the brackets used to infer the task.}
    \label{fig:encoder_ablation}
\end{figure}

We evaluate the three feature-extraction architectures GRU, MLP and Transformer on the \textit{half-cheetah-six} environment, i.e. omitting the \textit{jump} and \textit{front flip} tasks. We compare the three methods using 32 and 64 timesteps in the context.
The results are shown in Figure \ref{fig:encoder_ablation}. We can see that all methods show improved performance when using more timesteps in the context. As the GRU outperforms the other methods when 64 timesteps are used, we use this architecture in our study.

\subsection{Limitations}
The results of our experiments demonstrate that TIGR is applicable to broad and non-parametric environments with zero-shot adaptation to non-stationary task changes, where prior methods even fail with few-shot adaptation. 
However, our method is not applicable to sparse reward settings, since it assumes that the environment gives a feedback to the agent following a dense reward function.
In general, this drawback can presumably lead to weaker performance for problems that do not follow well-shaped reward functions, and we suppose that the \textit{front flip} task might also not be solved due to ill-defined rewards.

%% file: 07_conclusion.tex
\section{Conclusion} 
In this paper, we presented TIGR, an efficient meta-RL algorithm for solving non-parametric and non-stationary task distributions. 
Using our task representation learning strategy, TIGR is able to learn behaviors in non-parametric environments using zero-shot adaptation to non-stationary task changes.
Our encoder is based on a generative model represented as a VAE and trained by unsupervised MDP reconstruction. This makes it possible to capture the multi-modality of the non-parametric task distributions.
We report 3-10 times better sample efficiency and superior performance compared to prior methods on a series of environments including the novel non-parametric \textit{half-cheetah-eight} benchmark.

%% file: 08_appendix.tex
\section{Experiment Environments}
\label{appendix_env}

The OpenAI Gym toolkit~\cite{gym} provides many environments for RL setups that can be easily modified to meet our desired properties. 

\subsection{Half-cheetah-eight}

An environment that is often used in meta-RL is the \textit{half-cheetah} \cite{temporalconvolutions, modelbasedmeta, maml, promp, asynchronousdrl}, and therefore we have chosen it to demonstrate the performance of our proposed approach.
We provide a new benchmark consisting of eight non-parametric tasks requiring qualitatively distinct behavior as defined in Table \ref{tab:environments} and visualized in Figure \ref{fig:environments}.
Each environment contains internal parametric variability, in which the desired velocity or goal is sampled from a range of possible values.
%
Each task was verified individually to show that the correct behavior is learnt when a high return is achieved by the algorithm. The environments are pseudo-normalized such that the maximum possible reward is $0$ (i.e., when there is no deviation from the desired velocity/position), and the agent starts with a reward of $-1$ on each episode. We suggest that this is a very important feature of the environments, since the agent cannot distinguish tasks based on the magnitude of the reward alone. We assume that this increases the difficulty of the challenge as some kind of exploratory movement is required at the beginning of each episode to deduce what behavior is needed.

\begin{table*}[!t]
    \centering
    \caption{Non-parametric variability proposed for the \textit{half-cheetah-eight} environment.}
    \begin{tabular}{llrcll}
        \hline
        \textbf{Behaviour} & \textbf{Task Properties} & & & & \textbf{Objective} \\ \hline
        Run forward & Horizontal velocity & $1 \leq$ & $v_x^*$ & $\leq 5$ & \multirow{2}{*}{$r = -\abs{v_x^* - v_x}$} \\ \cline{0-4}
        Run backward & Horizontal velocity & $-5 \leq$ & $v_x^*$ & $\leq -1$ & \\ \hline
        Reach goal in front & Horizontal position & $5 \leq$ & $p_x^*$ & $\leq 25$ & \multirow{2}{*}{$r = -\abs{p_x^* - p_x}$} \\ \cline{0-4}
        Reach goal in back & Horizontal position & $-25 \leq$ & $p_x^*$ & $\leq -5$ & \\ \hline
        Front stand & Angular position & $\frac{\pi}{6} \leq$ & $p_y^*$ & $\leq \frac{\pi}{2}$ & \multirow{2}{*}{$r = -\abs{p_y^* - p_y}$} \\ \cline{0-4}
        Back stand & Angular position & $-\frac{\pi}{2} \leq$ & $p_y^*$ & $\leq -\frac{\pi}{6}$ & \\ \hline
        Front flip & Angular velocity & $2 \pi \leq$ & $v_y^*$ & $\leq 4 \pi$ & $r = -\abs{v_y^* - v_y}$ \\ \hline
        Jump & Vertical velocity & $1.5 \leq$ & $v_z^*$ & $\leq 3.0$ & $r = -\abs{v_z^* - \abs{v_z}}$ \\ \hline
    \end{tabular}
    \label{tab:environments}
\end{table*}

\subsection{Ant-three}

The second environment that is often used in meta-RL to demonstrate the generalization ability to multiple agents is the \textit{ant} \cite{pearl}. We modify the standard \textit{ant} environment and introduce three base tasks \textit{run}, \textit{reach goal} and \textit{jump}. \textit{Run} and \textit{reach goal} are divided into the four directions \textit{up}, \textit{down}, \textit{left} and \textit{right} but are considered as a single base task due to the ant's symmetricity. Task specifications are defined in Table \ref{tab:ant}.

\begin{table*}[!t]
    \centering
    \caption{Non-parametric variability proposed for the \textit{ant} environment.}
    \begin{tabular}{llrcll}
        \hline
        \textbf{Behaviour} & \textbf{Task Properties} & & & & \textbf{Objective} \\ \hline
        Run & Velocity \textit{up}, \textit{down}, \textit{left}, \textit{right} & $1 \leq$ & $v_{x/y}^*$ & $\leq 3$ & $r = -\abs{v_{x/y}^* - v_{x/y}}$ \\ \hline
        Reach goal & Position \textit{up}, \textit{down}, \textit{left}, \textit{right} & $5 \leq$ & $p_{x/y}^*$ & $\leq 15$ & $r = -\abs{p_{x/y}^* - p_{x/y}}$ \\ \hline
        Jump & Velocity & $0.5 \leq$ & $v_z^*$ & $\leq 2$ & $r = -\abs{v_z^* - v_z}$ \\ \hline
    \end{tabular}
    \label{tab:ant}
\end{table*}

\section{Evaluation Details}
\label{appendix_evaluation}
We carried out the experiments on an 32-core machine with 252GB of RAM and 8 Tesla V100 GPUs. We implemented TIGR in PyTorch (version 1.7.0) and ran it on Ubuntu 18.04 with Python 3.7.7. 
The implementation of TIGR is based on the PEARL implementation given by \cite{pearl}.
\begin{itemize}
    \item All curves in this work are plotted from three runs with random task initializations and seeds.
    \item Shaded regions indicate one standard deviation around the mean.
\end{itemize}

We give an overview of important hyperparameters of the method and the values we used during our experiments in Table \ref{tab:hyperparams_general}. The settings for the \textit{half-cheetah-eight} environment can be seen in Table \ref{tab:cheetah_hyperparams}. 
Detailed code can be found in the supplementary materials.

\begin{table*}[!t]
    \centering
    \caption{General hyperparameters.}
    \begin{tabular}{lr}
        \hline
        \textbf{Hyperparameter} & \textbf{Value}\\ \hline
        Optimizer & ADAM\\
        Learning rate encoder, decoder, SAC & 3e-4 \\
        Discount factor $\gamma$ & 0.99\\
        Entropy target $\mathcal{H}$ & $-\text{dim}(\mathcal{A})$\\
        SAC network size & $3 \times 300$ units\\
        Net complex $c_n$ & 5\\
        GRU input dim  & $\text{dim}(\mathcal{S}) + \text{dim}(\mathcal{A}) + 1 + \text{dim}(\mathcal{S}')$\\
        GRU hidden layer size & $c_n \times$ GRU input dim\\
        VAE network size & 2 layers [GRU hidden layer size \& \\
        & $\text{Num Classes} \times (2\times \text{Latent Dim} + 1)$]\\
        Dynamics network size & 2 layers [$c_n\times(\text{dim}(\mathcal{S}) + \text{dim}(\mathcal{A})  + \text{dim}(\mathcal{Z}))$] \\
        Reward network size & 3 layers [$c_n\times(\text{dim}(\mathcal{S}) + \text{dim}(\mathcal{A})  + \text{dim}(\mathcal{Z}))$] \\
        Non-linearity (all networks) & ReLU\\
        SAC target smoothing coefficient & 0.005\\
        Evaluation trajectories per task per epoch & 1\\
        Task inference training steps per epoch & 128\\
        Task inference training batch size & 4096\\
        Policy training steps per epoch & 2048\\
        Policy training batch size & 256\\
        Train-validation split encoder training & $0.8$ / $0.2$\\
        Loss weights: & \\
        - $\alpha_\text{KL-divergence}$ & $0.001$\\
        - $\beta_\text{euclid}$ & 5e-4\\
        - $\gamma_\text{classification}$ & $0.1$\\ \hline
    \end{tabular}
    \label{tab:hyperparams_general}
\end{table*}

\begin{table}[!t]
    \centering
    \caption{\textit{Half-cheetah-eight} hyperparameters.}
    \begin{tabular}{lr}
        \hline
        \textbf{Hyperparameter} & \textbf{Value}\\ \hline
        Training tasks & 80\\
        Test tasks & 40\\
        Maximal trajectory length & 200\\
        Training epochs & 2000\\
        Data collection: initial samples per task & 200\\
        Data collection: training tasks for sampling per epoch & 80\\
        Data collection: samples per task per epoch & 200\\
        Encoder context timesteps & 64\\
        Encoder latent dimension $\text{dim}(\bm z)$ & 8\\
        Encoder VAE components & 8\\ \hline
    \end{tabular}
    \label{tab:cheetah_hyperparams}
\end{table}


